%% file: arxiv.tex
\documentclass{article} %
\usepackage{iclr2027_conference,times}
\usepackage[T1]{fontenc} %

\input{math_commands.tex}

\usepackage{hyperref}
\usepackage{url}
\usepackage{enumitem}
\usepackage{tcolorbox}
\tcbuselibrary{breakable,skins}

\usepackage{graphicx}
\usepackage{booktabs}
\usepackage{array}
\usepackage{tabularx}
\usepackage{multirow}
\usepackage{makecell}
\usepackage{amsmath}
\usepackage{amssymb}
\usepackage{xcolor}
\usepackage{listings}
\usepackage{algorithm}
\usepackage{algpseudocode}
\newcommand{\belief}{\mathbf{b}}

\definecolor{yzBlue}{RGB}{59,130,246}

\definecolor{yzOrange}{HTML}{C2410C}

\hypersetup{
  colorlinks=true,
  linkcolor=yzOrange,
  citecolor=yzBlue,
  urlcolor=yzBlue
}

\title{EvoBelief: Falsifiable World Models for Self-Improving LLM Agents}

\author{
\makebox[0.30\textwidth][c]{\textbf{Janet Wang\textsuperscript{1$\ast\dagger$}}}
\makebox[0.30\textwidth][c]{\textbf{Jiang Guo\textsuperscript{2$\ast$}}}
\makebox[0.30\textwidth][c]{\textbf{Yi Xiang\textsuperscript{2}}}
\\
\makebox[0.30\textwidth][c]{\textbf{Jiaqing Yuan\textsuperscript{2}}}
\makebox[0.30\textwidth][c]{\textbf{Mingwen Dong\textsuperscript{2}}}
\makebox[0.30\textwidth][c]{\textbf{Hideo Kobayashi\textsuperscript{2}}}
\\
\makebox[0.30\textwidth][c]{\textbf{Shuaichen Chang\textsuperscript{2}}}
\makebox[0.30\textwidth][c]{\textbf{Chandan K. Reddy\textsuperscript{2}}}
\makebox[0.30\textwidth][c]{\textbf{Jiarong Jiang\textsuperscript{2}}}
\\[0.5em]
\makebox[0.90\textwidth][c]{%
\textsuperscript{1}Tulane University
\qquad\qquad
\textsuperscript{2}Amazon
}
}

\iclrfinalcopy %
\begin{document}

\maketitle
{\renewcommand{\thefootnote}{\fnsymbol{footnote}}%
\footnotetext[1]{Corresponding authors: \texttt{swang47@tulane.edu}, \texttt{gujiang@amazon.com}.}%
\footnotetext[2]{Work done during an internship at Amazon.}}
\suppressfloats[t] %
\lhead{Preprint.} %

\input{sections/abstract}
\input{sections/introduction}
\input{sections/related_work}
\input{sections/method}
\input{sections/experimental}
\input{sections/conclusion}

\bibliography{iclr2027_conference}
\bibliographystyle{iclr2027_conference}

\clearpage
\appendix
\input{sections/limitations}
\input{sections/appendix}

\end{document}

%% file: math_commands.tex
\usepackage{amsmath,amsfonts,bm}

\def\eqref#1{equation~\ref{#1}}

\def\1{\bm{1}}

\DeclareMathAlphabet{\mathsfit}{\encodingdefault}{\sfdefault}{m}{sl}
\SetMathAlphabet{\mathsfit}{bold}{\encodingdefault}{\sfdefault}{bx}{n}

%% file: sections/abstract.tex
\begin{abstract}
Self-evolving LLM agents typically improve how they act through accumulated experience, reflection, or skills. Yet even procedurally capable agents can fail when their beliefs about environment dynamics are incomplete, incorrect, or outdated. We study a complementary axis of self-evolution: evolving those beliefs during deployment. Such beliefs can support multiple tasks, but incorrect ones can also propagate across them. We introduce \textbf{EvoBelief}, a self-evolving world model that represents environment knowledge as falsifiable hypotheses. Unexpected outcomes and decision-relevant unknowns trigger hypothesis formation. Later observations and controlled experiments provide independent validation before a rule can guide predictions or plans within its supported scope. Counterexamples narrow or retire rules as the environment changes. The language model remains frozen, and each rule remains linked to its evidence and the decisions it changes. Across four benchmarks spanning static dynamics, hidden drift, and scientific discovery, EvoBelief improves task performance and reuses learned rules across tasks. On ALFWorld without actor-side transition memory, it improves success over the matched Frozen World Model by 23.9\% and 31.3\% with Claude Haiku 4.5 and GPT-5.4, respectively. On MiniGrid, it revises stale rules and recovers after unannounced dynamics changes.
In DiscoveryWorld's Reactor Lab, it discovers and validates quantitative laws, completing 83.3\% of Normal-tier and 58.3\% of Challenge-tier instances, versus 0\% for each evaluated baseline. 
These results demonstrate self-improvement through testable, revisable environment knowledge rather than model-parameter updates.
\end{abstract}

%% file: sections/introduction.tex
\section{Introduction}

\begin{figure}[t]
    \centering
    \includegraphics[width=\linewidth]{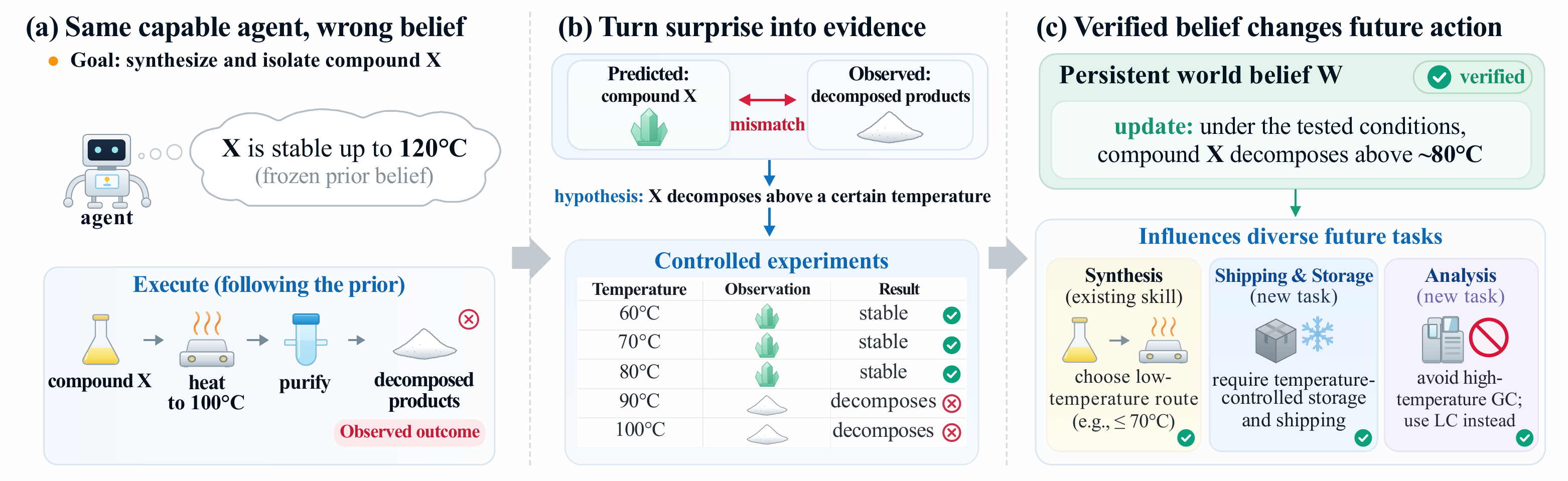}
    \caption{Illustrative example. (a) Following its frozen prior, a capable agent heats compound X to 100\,$^{\circ}$C and gets decomposition products. (b) The mismatch triggers a hypothesis, which controlled experiments test. (c) The verified belief is stored in the persistent world model $W$ and changes later actions in the original task and in new ones, without any parameter update. Based on the updated belief, the agent avoids gas chromatography (GC), whose operating temperature exceeds the supported range, and correctly chooses to use liquid chromatography (LC) instead.}
    \label{fig:evobelief-teaser}
\end{figure}

Existing self-evolving LLM agents mostly adapt to new or changing environments by improving \emph{how they act}. They learn from past interaction trajectories, refine behavior through reflection, accumulate memories, or distill experience into reusable skills \citep{zhao2023expel,shinn2023reflexion,park2023generative,wang2023voyager}.
These mechanisms are effective in addressing failures arising from inadequate strategy or procedure.
But an agent also acts on its prior beliefs about how the environment works, and in a new environment these beliefs can be wrong. Consider the agent in Figure~\ref{fig:evobelief-teaser}, whose task is to synthesize a compound X. It already knows how to heat, purify, and analyze samples, yet it fails because its prior predicts that X is stable up to 120\,$^{\circ}$C, while X in fact decomposes at 100\,$^{\circ}$C. A task-specific correction, such as ``use a lower temperature,'' can address the immediate failure but does not capture the underlying property. The agent should instead test how X's stability depends on temperature and retain the resulting evidence-supported constraint. This belief can then guide not only synthesis, but also related tasks such as shipping, storage, and analysis.

This motivates \textbf{belief evolution}, a complementary form of self-evolution that updates a persistent model of environment dynamics rather than only adapting task-specific behavior.
A policy specifies how an agent acts toward a goal, $\pi(a \mid s,g)$, whereas a world model captures how the environment responds, $W(s' \mid s,a)$. When the world model is wrong, even a capable agent may choose actions based on incorrect predicted outcomes. Conversely, correcting an environmental belief can improve decisions across multiple tasks without separately updating a procedure or skill for each task. This raises a natural question: \emph{how can an LLM agent form, test, revise, and reuse beliefs about its environment while keeping the underlying language model frozen?}

A growing line of work has begun to adapt world models for LLM agents. WALL-E and WALL-E 2.0 induce rules from collected trajectories to align an LLM world model with the environment, and keep the rules that correct the most prediction errors in the collected trajectories \citep{zhou2024walle,zhou2025walle2}. Imagine-then-plan approaches such as ITP-I use offline-trained world models to simulate future outcomes during planning, while keeping the world model itself fixed \citep{liu2026itp}. WorldEvolver is closest to our setting, updating world-model knowledge online from prediction mismatches with a memory of past episodic transitions and semantic facts \citep{zhang2026worldevolver}.
These works demonstrate the value of environment-specific world models for agents, but leave open three challenges central to continual adaptation: \textbf{validation}, \textbf{maintenance}, and \textbf{application}. Newly inferred rules must first be validated against additional evidence before being trusted, and once established, they must be maintained as the environment changes, by being revised, narrowed, or retired when contradicted by new evidence. Finally, validated beliefs should influence future predictions and decisions, including tasks different from the one in which they were acquired.

\begin{figure}[t]
    \centering
    \includegraphics[width=0.82\linewidth]{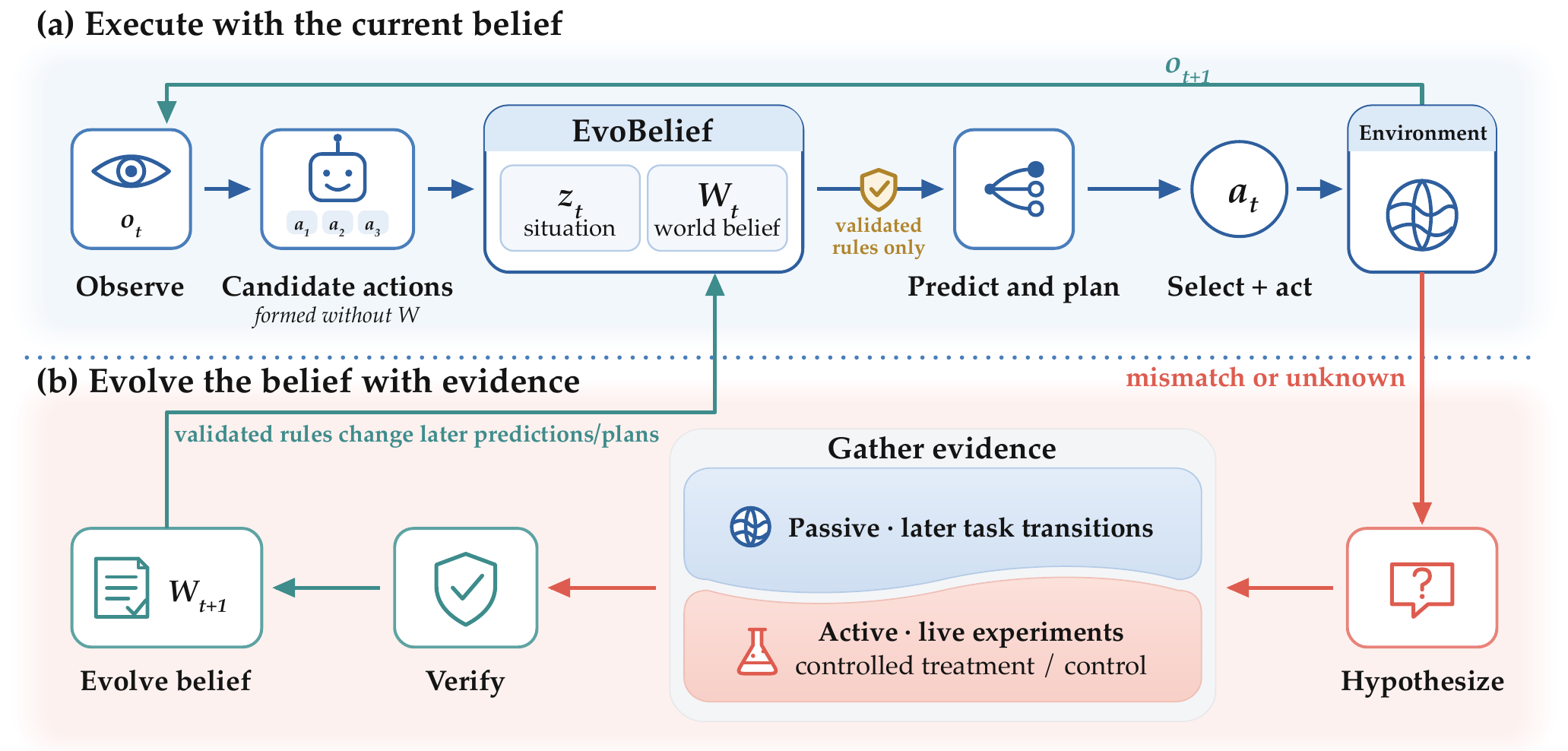}
    \caption{\textbf{Verified belief revision closes the loop from experience to later behavior.} EvoBelief uses the current $z_t$ and $W_t$ to predict action consequences. A grounded prediction mismatch or a decision-critical unknown opens an evidence loop that forms falsifiable hypotheses, gathers evidence, and verifies their scope. Only validated, applicable rules guide predictions and plans.}
    \label{fig:evobelief-overview}
\end{figure}

We introduce \textbf{EvoBelief}, a self-evolving world model that addresses these challenges through an explicit cycle of hypothesis formation, validation, and revision (Figure~\ref{fig:evobelief-overview}). When an outcome contradicts the model's prediction, or when an unknown matters for a decision, EvoBelief proposes falsifiable hypotheses about the local environment dynamics involved. It tests these hypotheses against later observations or through controlled experiments, promotes evidence-supported hypotheses into reusable rules, and revises, narrows, or retires rules when later evidence contradicts them. Only validated rules are allowed to influence the agent's future predictions and plans.

EvoBelief maintains a two-timescale belief representation, capturing episode-local situational state ($z_t$) and persistent, evidence-linked beliefs about environment dynamics ($W_t$). The underlying LLM remains frozen, and all adaptation occurs in this external world model, separating the model's fixed parametric prior from an environment-specific belief that evolves through test-time interaction.
This explicit representation also makes belief evolution inspectable. Each rule is linked to the evidence that supports it, the conditions under which it applies, and subsequent observations that may challenge it. This design allows us to trace how a validated rule guided a prediction or plan, and evaluate how learned beliefs transfer across tasks, influence decisions, and adapt when the environment changes. Our contributions are:

\begin{itemize}
[leftmargin=*, nosep] %
    \item We introduce \textbf{belief evolution} as a complementary form of self-improvement for LLM agents, in which a persistent world model captures environment-specific beliefs that can evolve at test time through interaction and generalize across tasks without updating the underlying language model.
    \item We develop a belief lifecycle for \textbf{validation}, \textbf{maintenance}, and \textbf{application}: hypotheses are validated before being trusted, established rules can be revised or retired as evidence changes, and validated beliefs are applied to relevant future predictions and decisions.
    \item We evaluate EvoBelief on four benchmarks covering static and changing dynamics, and scientific discovery, showing improved task success, cross-task transfer of learned beliefs, adaptation to shifts in dynamics, and discovery of quantitative laws through active experimentation.

\end{itemize}

%% file: sections/related_work.tex
\section{Related Work}
\label{app:related}

\textbf{Belief states and world models.}
In a POMDP, the belief state is a distribution over latent environment states \citep{kaelbling1998planning}, and predictive state representations characterize state through predictions of future observations \citep{littman2001predictive}. World models address the complementary question of how actions change the world, from planning with learned dynamics to recurrent and latent models \citep{sutton1990integrated,ha2018world,hafner2020dream,assran2025vjepa2,wang2026adajepa}. Recent language-agent work externalizes state uncertainty or maintains linguistic beliefs under partial observability \citep{singh2026agentbrace,murphy2026blf,jiang2026pabu,liao2026beliefmemory}, and the belief-based world model of \citet{kumar2026bbwm} exposes a hand-specified belief over the current state to the policy but does not learn it from interaction. EvoBelief separates the two roles across timescales: $z_t$ estimates the current situation, and $W_t$ is a persistent belief over local environment dynamics learned from interaction. Related world-model agents plan with the model as a simulator, inject world knowledge into planning, learn environment rules from experience, or evolve transition memories \citep{hao2023rap,qiao2024wkm,zhou2024walle,zhou2025walle2,liu2026itp,fang2025webevolver,zhang2026worldevolver}; others explore curiosity-driven exploration, deliberate environment probing, theory formation, and closed-loop scientific discovery \citep{pathak2017curiosity,levy2025worldllm,song2026envprobe,kabra2026autoscilab,jin2026htr}, and interactive benchmarks now require such discovery \citep{wang2022scienceworld,jansen2024discoveryworld}. EvoBelief differs in that only validated, scoped claims may change a prediction or a plan.

\textbf{Experience, memory, and self-evolving agents.}
LLM agents improve from experience through reflections, episodic memories, reusable lessons, procedures, and skills \citep{shinn2023reflexion,zhao2023expel,park2023generative,wang2023voyager,ouyang2025reasoningbank,ouyang2026skillos,chen2024automanual,ma2026proplay,shang2026hypothesis}, and a broader self-evolving literature modifies prompts, scaffolds, search procedures, or agent code while keeping the model fixed \citep{he2026evotest,li2026jitrl,zhang2026dgm,robeyns2025sica,lee2026metaharness,xu2026harness}. ProPlay learns procedural world models, HDSO optimizes skills through hypothesis-driven testing, and AutoManual builds environment instruction manuals from interaction \citep{ma2026proplay,shang2026hypothesis,chen2024automanual}. EvoBelief instead maintains predictive claims about environment dynamics with explicit conditions, evidence-supported scope, and counterexample-driven revision. These approaches primarily change \emph{how the agent acts}; EvoBelief instead changes \emph{what the agent believes about the deployed environment} and separates belief existence from behavioral authority, which a learned claim gains only after independent validation and can lose to later counterexamples. That lifecycle is also what continual adaptation asks for when transition dynamics change and earlier knowledge goes stale \citep{xie2021continual,wolczyk2021continualworld,kessler2023effectiveness}. Belief-calibrated optimization is closest in vocabulary \citep{chen2026bco}, but its beliefs concern how an evaluator responds to edits of the agent's own scaffold rather than the deployed environment's dynamics.

%% file: sections/method.tex
\section{Method}
\label{sec:method}

\subsection{Problem Formulation}
\label{sec:setting}
At time $t$, an environment has latent state $s_t\in\mathcal{S}$, emits an observation $o_t\in\mathcal{O}$, and transitions after action $a_t\in\mathcal{A}$. A task is specified by goal $g$, and the agent observes the interaction history $h_t=(o_0,a_0,o_1,\ldots,a_{t-1},o_t)$. The environment may contain local dynamics that disagree with the frozen language model's pretrained prior, and those dynamics may change during deployment. With the model parameters $\theta$ fixed, EvoBelief maintains two externalized states,
\begin{equation}
    \belief_t=(z_t,W_t).
\end{equation}
Here $z_t$ answers the local question of what the agent currently takes to be true about this situation. It records observed entities, relations, quantities, and task progress, and is reset at each episode boundary. $W_t$ answers the persistent question of how this environment responds to actions. It survives episode resets and changes only as new evidence is incorporated. The frozen LLM therefore supplies a general prior, while $W_t$ stores environment-specific knowledge learned during deployment.

\subsection{World Knowledge as Testable Hypotheses}
\label{sec:object}
\label{sec:formation}

The basic unit of knowledge in $W_t$ is a falsifiable \emph{hypothesis} $H$ about an observable action consequence. We write its claim as
\begin{equation}
    H \mathrel{:=} (C \wedge A \Rightarrow Y),
    \label{eq:hypothesis}
\end{equation}
which states that when condition $C$ holds, taking action or procedure $A$ should produce observable consequence $Y$. This form makes the claim directly testable, since one can establish $C$, execute $A$, and check whether $Y$ occurs. Each hypothesis also stores the scope in which it is supported, a recipe for testing it, and an evidence record. $A$ may be a single action, a multi-step procedure, or a measurement intervention, while $Y$ may be a state change or a measured quantity.

A hypothesis may contain both fixed entities and abstract object roles. When it is used, those roles are \textbf{bound} to concrete objects (i.e. \textbf{bindings}) in the current context, and broader scope is earned only from evidence across varied bindings rather than assumed in advance. For example, the claim that heating above about 80\,$^{\circ}$C decomposes the sample is first bound to compound X, and it covers other compounds only after evidence from them.

Hypotheses are formed from the agent's own interaction, starting from an empty $W_0$. After an action, EvoBelief compares the realized outcome with the prediction made before acting. A grounded \emph{prediction mismatch} suggests that the current model of the environment is wrong, and a \emph{decision-critical unknown}, one whose resolution would change what the agent should do next, suggests that it is incomplete. Because one surprising outcome can have several explanations, EvoBelief may keep \emph{rival hypotheses} until later evidence distinguishes them. In the compound X example, decomposition could depend on temperature, heating duration, or the vessel. Forming a hypothesis is not evidence for it. An unvalidated hypothesis cannot justify a task-directed action as established knowledge, but it may motivate an experiment. Details on hypothesis identity and admissible evidence are in Appendix~\ref{app:method-details}.

\subsection{Testing Hypotheses with Passive and Active Evidence}
\label{sec:testing}

Evidence can arrive in two ways. \emph{Passive evidence} comes from an ordinary task transition that happens to test an existing hypothesis, as when a later task heats compound X and records whether it remains intact. \emph{Active evidence} comes from an \emph{experiment} run specifically to resolve an uncertainty or distinguish among rival hypotheses. Both routes update the same evidence record. Active experiments use live environment actions and therefore consume the task's interaction budget. EvoBelief admits one only when its outcome can change a decision the agent must make, no validated rule already resolves the question, and the required setup is executable. Before the experiment runs, its support and falsification criteria are fixed. The realized outcome then settles the test as \emph{supported}, \emph{counterexampled}, or \emph{inconclusive}. An inconclusive test means that the required condition was not actually witnessed, or that the outcome was missing or ambiguous, and it counts as neither support nor counterevidence. Unresolved hypotheses remain pending and can be tested later. Experiment admission and counting details are in Appendix~\ref{app:method-details}.

\subsection{From Evidence to Authority: The Rule Lifecycle}
\label{sec:lifecycle}

Evidence determines what a hypothesis is allowed to do. We use \emph{authority} to mean permission for a learned claim to guide a later prediction or plan. A newly formed hypothesis has none. Permission to reuse must be earned from independent evidence across episodes, bindings, and tests. For example, one batch of compound X that decomposed at 90\,$^{\circ}$C is not enough, and the claim must hold again in a later episode, on a different batch, before it may guide heating decisions.

Let $S$ be the supports of a hypothesis, and let $N_{\mathrm{ep}}(S)$, $N_{\mathrm{bind}}(S)$, and $N_{\mathrm{test}}(S)$ count the distinct episodes, bindings, and independent tests represented in those supports. A hypothesis is committed to $W_t$ as a \emph{validated rule} only when
\begin{equation}
\mathrm{Commit}(H)
\Longleftrightarrow
\bigl(N_{\mathrm{ep}}(S)\ge\tau_{\mathrm{ep}}\bigr)
\wedge
\bigl(N_{\mathrm{bind}}(S)\ge\tau_{\mathrm{bind}}\bigr)
\wedge
\bigl(N_{\mathrm{test}}(S)\ge\tau_{\mathrm{test}}\bigr).
\label{eq:commit}
\end{equation}
The thresholds are set before any benchmark run, are never tuned on benchmark results, and are listed in Appendix~\ref{app:method-details}. Each generalized role must vary across supporting bindings, and the event that first formed a hypothesis can never by itself grant reusable authority. 
Counterevidence can later remove that authority. EvoBelief first narrows the rule to the scope still supported by its surviving evidence and recomputes the commitment condition. If the surviving evidence no longer satisfies Eq.~\ref{eq:commit}, the rule is \emph{demoted}. It loses its reusable authority but keeps its evidence, so later support can restore it. A rule is \emph{retired} and no longer used only when repeated counterexamples leave it with no supporting evidence. Because the evidence record is preserved, every change in status can be traced to the observations that caused it. The same lifecycle handles drift in environment dynamics. A previously useful rule can lose authority under changed dynamics, while new evidence supports a successor.

\subsection{Using the World Model for Prediction and Planning}
\label{sec:consumption}
A validated rule matters only when it applies to the current situation. At each decision, the agent first forms candidate actions $A_t$ without access to $W_t$. For each candidate $a\in A_t$, the frozen predictor supplies an anchor prediction
\begin{equation}
    p_t^0(a)=F_\theta(a,g,z_t),
\end{equation}
which captures what the agent would expect without self-learned world knowledge.

EvoBelief then retrieves validated rules relevant to the candidate and checks whether their conditions hold and whether the current binding lies inside their evidence-supported scope. This check uses the state estimate $z_t$ when assessing candidates and the state predicted after each earlier step when working out a sequence of actions. Retrieval alone has no effect. Only an applicable validated rule may change the anchor prediction.

Those rule-supported consequences affect behavior in two ways. First, they can revise the expected outcome of a candidate action, changing its ranking or ruling it out when a required precondition is violated. For example, once the decomposition rule for compound X is validated, it changes the predicted outcome of heating X to 100\,$^{\circ}$C, so that candidate falls below a lower-temperature route. A rule whose condition does not hold makes no prediction, so it cannot rule out an action by itself. Second, compatible rules can be chained into a short plan when the consequence of one rule establishes the condition of another, and the agent can execute the plan's first step even if it is not among the candidates in $A_t$. A validated quantitative relation can likewise set the parameter values needed to obtain a target consequence (for example, heating compound X below 80\,$^{\circ}$C, where it stays intact).

If no validated rule applies, EvoBelief falls back to the same $W$-blind choice it would otherwise have made. If an unresolved claim is decision-relevant, an admissible experiment may instead be selected to test it (\S\ref{sec:testing}). Whenever a validated rule changes an executed action, EvoBelief records the rule, the prediction or plan it changed, and the realized outcome. A success caused by using the rule is recorded for analysis but does not count as new independent support for the same rule. Appendix~\ref{app:prompts} gives the benchmark-specific interfaces and the model-call inventory.

%% file: sections/experimental.tex
\section{Experiments}
\label{sec:experimental}

\begin{table}[t]
\centering
\small
\setlength{\tabcolsep}{4.5pt}
\caption{Success rate (\%) with and without actor-side transition memory, on ALFWorld and ScienceWorld for both backbones. All methods are run through the same harness, interface, and budgets.}
\label{tab:backbone-results}
\begin{tabular}{llcccc}
\toprule
& & \multicolumn{2}{c}{ALFWorld} & \multicolumn{2}{c}{ScienceWorld} \\
\cmidrule(lr){3-4}\cmidrule(lr){5-6}
Setting & Method & Haiku 4.5 & GPT-5.4 & Haiku 4.5 & GPT-5.4 \\
\midrule
\multirow{5}{*}{No memory}
& ReAct & 33.58 & 46.27 & 55.56 & 41.11 \\
& ITP-I \citep{liu2026itp} & 29.85 & 39.55 & 42.22 & 36.67 \\
& WorldEvolver \citep{zhang2026worldevolver} & 35.07 & 61.94 & 52.22 & 65.56 \\
& Frozen WM & 42.54 & 44.78 & 46.67 & 53.33 \\
& EvoBelief & \textbf{66.42} & \textbf{76.12} & \textbf{64.44} & \textbf{71.11} \\
\midrule
\multirow{5}{*}{Memory}
& ReAct & 45.52 & 50.00 & 50.00 & 46.67 \\
& ITP-I \citep{liu2026itp} & 35.07 & 40.30 & 47.78 & 42.22 \\
& WorldEvolver \citep{zhang2026worldevolver} & 58.96 & 72.39 & 64.44 & 67.78 \\
& Frozen WM & 57.46 & 55.22 & 47.78 & 58.89 \\
& EvoBelief & \textbf{68.66} & \textbf{76.87} & \textbf{68.89} & \textbf{72.22} \\
\bottomrule
\end{tabular}
\end{table}

\subsection{Setup}
\label{sec:setup}

We evaluate on four benchmarks spanning three deployment regimes: ScienceWorld and ALFWorld (static, largely prior-aligned dynamics) \citep{wang2022scienceworld, shridhar2021alfworld}, MiniGrid (hidden dynamics drift) \citep{chevalier2023minigrid}, and DiscoveryWorld (scientific discovery) \citep{jansen2024discoveryworld}. Splits, budgets, and configurations are in Appendix~\ref{app:setup-details}. We compare against ReAct, WorldEvolver, and \emph{Frozen WM}. Frozen WM uses the same $W$-blind actor and predictor as EvoBelief but lets no learned rule guide a prediction or a plan. The comparison with Frozen WM evaluates the complete EvoBelief system relative to this matched prior-driven control. ALFWorld and ScienceWorld are evaluated without and with actor-side transition memory, provided identically to every method. EvoBelief runs without actor-side transition memory on MiniGrid and DiscoveryWorld. Claude Haiku 4.5 and  GPT-5.4 are the backbones used.
Within each comparison all methods share the benchmark interface, task order, seeds, and budgets, and each task is attempted once. 

\paragraph{Implementation.} On ALFWorld, MiniGrid, and ScienceWorld, action--effect rules are induced from typed transitions and applied to predict state changes for planning and veto, without a rule-conditioned model call. On DiscoveryWorld, quantitative and categorical laws are fitted to measurements and observations, and validated by a held-out prediction frozen before the intervention. Only DiscoveryWorld's frozen predictor engine, the fixed prediction component shared with Frozen WM, makes rule-conditioned model calls.
EvoBelief uses the same role-level model interfaces on every benchmark. Its representations and control components are fixed before evaluation (Appendix~\ref{app:prompts}).

\subsection{Regime I: Static, Largely Prior-Aligned Environments}
\label{sec:results-static}

ScienceWorld and ALFWorld test whether belief evolution helps when the environment is stationary and much of its dynamics remain compatible with the pretrained prior. Table~\ref{tab:backbone-results}, which also includes the world-model lookahead baseline ITP-I, shows that EvoBelief has the highest task success rate in every setting shown, but the route by which the belief layer helps differs across the two environments.

\begin{figure}[t]
    \centering
    \includegraphics[width=\linewidth]{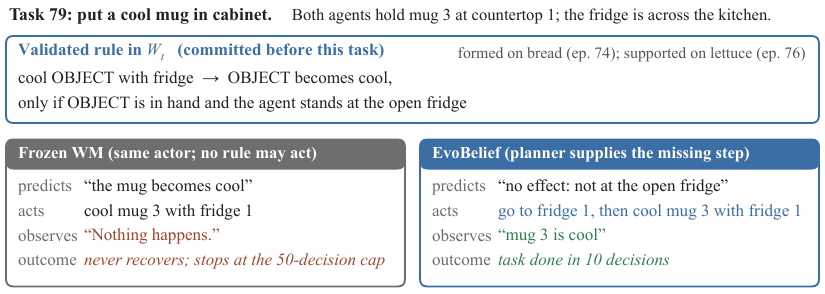}
    \caption{A rule-induced decision change on a real ALFWorld trace (Haiku 4.5, unseen task). Frozen WM predicts that cooling succeeds, executes the draft to no effect, and never recovers. EvoBelief uses a validated cooling rule, moves to the fridge first, and completes the task in ten decisions.}
    \label{fig:alfworld-flip}
\end{figure}

On ALFWorld, persistent validated rules are the main behavioral route. They correct local action mechanics that the prior gets wrong or leaves underspecified, supply missing prerequisites, and prevent the agent from repeating failed procedures. In the trace of Figure~\ref{fig:alfworld-flip}, for example, the learned cooling rule changes the predicted consequence of cooling, inserts the missing move to the fridge, and completes the task.

\begin{figure}[t]
    \centering
    \includegraphics[width=0.215\textwidth]{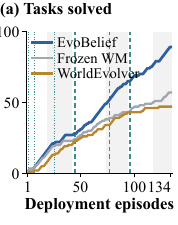}\hfill
    \includegraphics[width=0.215\textwidth]{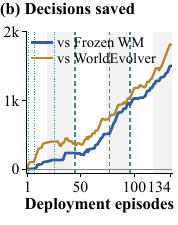}\hfill
    \includegraphics[width=0.275\textwidth]{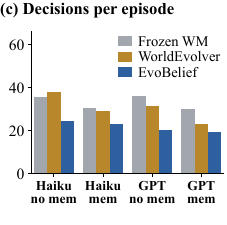}\hfill
    \includegraphics[width=0.275\textwidth]{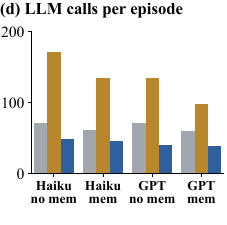}
    \caption{\textbf{Learning over deployment and efficiency on ALFWorld.} (a) Tasks solved so far over the 134 unseen tasks (Haiku 4.5, no transition memory). Shaded blocks are the six task types, and dotted and dashed lines mark when the generic and the clean, cool, and heat rules were committed. (b) Decisions saved by EvoBelief over Frozen WM and WorldEvolver. (c, d) Mean decisions and model calls per episode for both backbones, without and with memory (lower is better).}
    \label{fig:alfworld-deployment}
\end{figure}

Because $W_t$ persists across tasks, generic household rules learned early are reused across later task families, while process-specific rules such as clean, heat, and cool are reused across new object bindings within their supported family (Tables~\ref{tab:alfworld-transfer} and~\ref{tab:alfworld-transfer-rules}). Across the four ALFWorld configurations, over 80\% of the decisions under a validated rule fall in a family other than the one where the rule was validated (Table~\ref{tab:alfworld-transfer}). This reuse also reduces repeated exploration, yielding more successes with fewer decisions and model calls (Figure~\ref{fig:alfworld-deployment}). On ALFWorld, EvoBelief's margin over Frozen WM is largest without transition memory for both backbones and remains positive with it, consistent with episodic recall and explicit world rules supplying partly overlapping cross-episode information.

ScienceWorld exercises the same belief lifecycle but through a different route. Live experiments account for most direct behavioral intervention, while validated rules guide a smaller number of later decisions. The experiments resolve decision-relevant unknowns using the same evidence and validation machinery that stores persistent rules, and Table~\ref{tab:mechanism-summary} summarizes the resulting experiment steps and rule-induced decisions. Thus the two static environments test complementary uses of the same world belief, persistent rule reuse across later tasks in ALFWorld and targeted evidence gathering with more selective rule consumption in ScienceWorld. Family breakdowns and paired tests are in Appendix~\ref{app:additional-results}.

\subsection{Regime II: Hidden Dynamics Drift}
\label{sec:results-dynamic}

\begin{figure}[t]
    \centering
    \includegraphics[width=0.2235\textwidth]{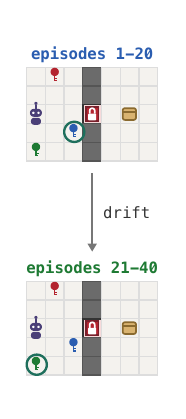}\hfill%
    \includegraphics[width=0.7576\textwidth]{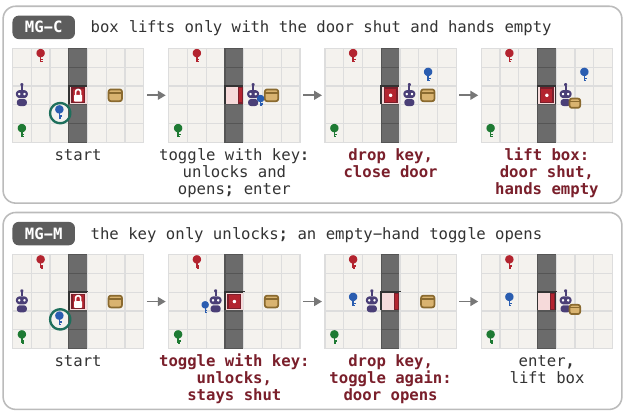}
    \caption{\textbf{The two central MiniGrid drift families.} Each episode is a two-room grid with a locked door, three keys, and a target box in the far room. \textbf{Left:} the drift shared by every family. Only the ringed key unlocks the door, blue for episodes 1--20 and green for 21--40. The switch is silent, and the red key that matches the door is a decoy. \textbf{Right:} the preconditions each family requires to complete the task of lifting the box. In MG-C the box lifts only with the door shut and hands empty. In MG-M the toggle only unlocks the door, and a second, empty-handed toggle opens it. Dark red marks steps the pretrained prior does not expect. Figure~\ref{fig:minigrid-storyboard} in the appendix shows all three families.}
    \label{fig:minigrid-env}
\end{figure}

\begin{table}[t]
\centering
\small
\setlength{\tabcolsep}{8pt}
\caption{MiniGrid task success (\%) before and after the hidden key switch. Each cell pools 60 episodes from three 40-episode streams per family; MG-C and MG-M use development seeds and MG-R uses fresh control seeds. Haiku 4.5, without transition memory. Higher is better.}
\label{tab:minigrid-drift}
\begin{tabular}{@{}lrrrrrr@{}}
\toprule
& \multicolumn{2}{c}{MG-C} & \multicolumn{2}{c}{MG-M} & \multicolumn{2}{c}{MG-R} \\
\cmidrule(lr){2-3}\cmidrule(lr){4-5}\cmidrule(lr){6-7}
Method & Pre-drift & Post-drift & Pre-drift & Post-drift & Pre-drift & Post-drift \\
\midrule
Frozen WM & 0.00 & 0.00 & 1.67 & 0.00 & 48.33 & 50.00 \\
WorldEvolver & 0.00 & 0.00 & 6.67 & 16.67 & 20.00 & 26.67 \\
\textbf{EvoBelief} & \textbf{61.67} & \textbf{95.00} & \textbf{78.33} & \textbf{81.67} & \textbf{83.33} & \textbf{78.33} \\
\bottomrule
\end{tabular}
\end{table}

\begin{figure}[t]
    \centering
    \includegraphics[width=0.48\textwidth]{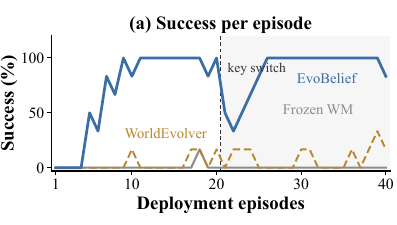}\hfill
    \includegraphics[width=0.48\textwidth]{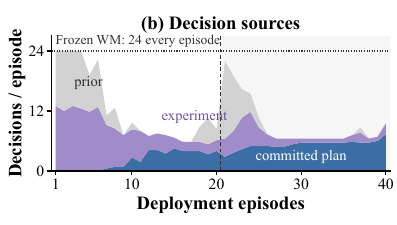}
    \caption{Adaptation over deployment on MiniGrid. (a) Success by episode. The dashed line marks the silent key switch after episode 20. (b) EvoBelief decisions per episode by source: pretrained prior, active experiment, or committed plan. Both panels average the six development streams from MG-C and MG-M.}
    \label{fig:minigrid-deployment}
\end{figure}

MiniGrid tests whether a learned world model stays useful when the environment changes silently during deployment. In every 40-episode stream the key that opens the door switches after episode 20, and MG-C and MG-M add preconditions the prior does not expect (Figure~\ref{fig:minigrid-env}). EvoBelief solves most episodes before and after the switch (Table~\ref{tab:minigrid-drift}), while Frozen WM and WorldEvolver rarely solve MG-C or MG-M, where success requires the hidden rule. On MG-R, a less adversarial control where the prior solves about half of the episodes, EvoBelief still improves substantially, so the gain is not explained by a weak prior or a floor effect. Over the deployment (Figure~\ref{fig:minigrid-deployment}), early episodes go to experiments on uncertain dynamics, after which validated rules carry short plans reused across episodes. When the silent key switch makes the door rule fail, the counterevidence retires the rule and reopens experimentation. EvoBelief then validates a successor rule, and success returns to its pre-switch level. On three held-out seeds EvoBelief solves 69.6\% of episodes against 1.3\% for Frozen WM, and 80.8\% against 7.1\% when every color and room name is permuted, so the rules are grounded in observed effects rather than entity names (Table~\ref{tab:minigrid-heldout}). Mechanism ablations (Table~\ref{tab:minigrid-ablation}) show that both sides of the loop matter. Without active experiments the critical rules never gain independent support, and without learned rules the planner falls back to the prior, so both variants stay at the prior's level. Keeping validated rules in shadow, with the same experiments but no executed plan step, also lowers success, most on MG-C, whose solution is a multi-step chain. Lifecycle traces and the remaining ablations are available in Appendix~\ref{app:drift-details}.

\subsection{Regime III: Scientific Discovery}
\label{sec:results-discovery}

\begin{figure}[t]
    \centering
    \includegraphics{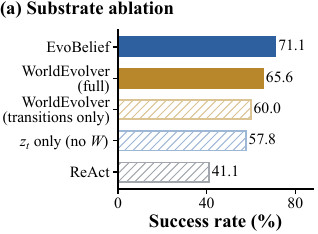}\hfill
    \includegraphics{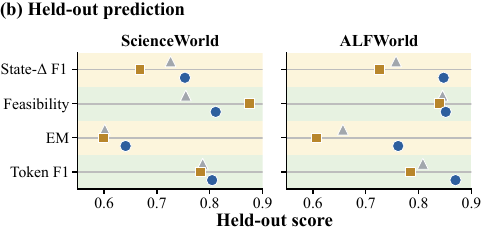}\\[1pt]
    \includegraphics{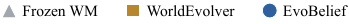}
    \caption{\textbf{Substrate ablation and held-out prediction.} (a) Success on the ScienceWorld tasks of Table~\ref{tab:backbone-results} with GPT-5.4 and no transition memory, by persistent substrate. Hatched bars store no explicit rules. (b) Structured state-change prediction on 1{,}500 held-out transitions per benchmark with Claude Haiku 4.5. State-$\Delta$ F1 measures how well the predicted state changes match the realized ones, Feasibility whether the action is correctly predicted to execute, and exact match (EM) and token F1 compare the predicted and realized state records. Exact prediction scores are in Table~\ref{tab:scienceworld_prediction}.}
    \label{fig:analysis}
\end{figure}

\begin{table}[t]
\centering
\small
\setlength{\tabcolsep}{4pt}
\caption{DiscoveryWorld results with GPT-5.4 on twelve fresh task instances per task family and difficulty tier. Completion, Knowledge, and Procedural are the benchmark's task-completion, blind knowledge, and subgoal-progress scores, in percent.}
\label{tab:discoveryworld-results}
\begin{tabular}{@{}lrrrrrr@{}}
\toprule
& \multicolumn{3}{c}{Normal} & \multicolumn{3}{c}{Challenge} \\
\cmidrule(lr){2-4} \cmidrule(lr){5-7}
Method & Completion & Knowledge & Procedural & Completion & Knowledge & Procedural \\
\midrule
\multicolumn{7}{@{}l}{\textit{Reactor Lab}} \\
ReAct & 0.00 & 0.00 & 70.33 & 0.00 & 0.00 & 64.66 \\
WorldEvolver & 0.00 & 0.00 & 72.67 & 0.00 & 0.00 & 69.54 \\
Hypothesizer & 0.00 & 0.00 & 74.00 & 0.00 & 0.00 & 68.39 \\
\textbf{EvoBelief} & \textbf{83.33} & \textbf{79.17} & \textbf{89.00} & \textbf{58.33} & \textbf{62.50} & \textbf{84.20} \\
\midrule
\multicolumn{7}{@{}l}{\textit{Plant Nutrients}} \\
ReAct & 41.67 & 16.67 & 76.85 & 50.00 & 0.00 & 77.78 \\
WorldEvolver & 16.67 & 8.33 & 70.37 & 75.00 & 0.00 & 83.33 \\
Hypothesizer & 50.00 & 0.00 & 79.63 & 75.00 & 0.00 & 84.26 \\
\textbf{EvoBelief} & \textbf{75.00} & \textbf{75.00} & \textbf{82.41} & \textbf{83.33} & \textbf{33.33} & \textbf{85.19} \\
\bottomrule
\end{tabular}
\end{table}

DiscoveryWorld tests whether EvoBelief can acquire environment laws that neither prior knowledge nor episodic recall can supply. We evaluate two themes at the Normal and Challenge tiers. Reactor Lab hides a quantitative relation between a measured crystal property and the reactor's resonance, so the target resonance must be computed from a law inferred from the agent's own measurements. Plant Nutrients hides a categorical rule for which nutrient settings make a planted seed grow.

In Reactor Lab, no baseline (including the benchmark's native Hypothesizer) completes a task instance or receives knowledge credit, whereas EvoBelief completes most Normal and a majority of Challenge task instances (Table~\ref{tab:discoveryworld-results}). Every completed Reactor Lab task instance follows the intended lifecycle (Tables~\ref{tab:dw-mechanism} and~\ref{tab:dw-laws}), in which rival quantitative laws are tested, the survivor is validated by a prediction frozen before a held-out crystal is tried, and the validated law configures the target reactor.

Plant Nutrients tests the same loop with a categorical rule whose consequence is realized only through a multi-step planting-and-growth procedure. 
EvoBelief completes the most task instances in both tiers and is the only method that earns knowledge credit in both. 
The reported confidence interval for its knowledge-score difference from each baseline excludes zero, while the interval for its completion difference from the strongest baseline of each tier includes zero (Table~\ref{tab:paired-stats}), so discovering the rule and completing the downstream task are related but distinct challenges. Per-instance laws, the paired knowledge margins, failure cases, and the development protocol are in Appendix~\ref{app:dw-details}.

\subsection{Ablation and Analysis}

\label{sec:analysis}

To understand what drives these gains, we examine which persistent substrate helps most, whether the learned belief predicts held-out transitions, and how it changes behavior in each environment. In Figure~\ref{fig:analysis} (a), explicit falsifiable rules outperform WorldEvolver with transition memory alone or with its full memory substrate, as well as the $z_t$-only variant and ReAct. In Figure~\ref{fig:analysis} (b), EvoBelief leads on State-$\Delta$ F1, exact match, and token F1 on both benchmarks, while WorldEvolver has higher feasibility accuracy on ScienceWorld but lower content scores. The behavioral route differs by environment (Table~\ref{tab:mechanism-summary}). Passively acquired rules drive planning and veto across tasks on ALFWorld, validated plans dominate on MiniGrid, where retirements track the hidden switch, and validated laws set plan parameters on DiscoveryWorld. ScienceWorld records many live experiment steps but comparatively few decisions guided by validated rules, so most direct behavioral intervention there comes from targeted experiments. Gap sizes and design notes are in Appendix~\ref{app:additional-results}.

%% file: sections/conclusion.tex
\section{Conclusion \& Discussion}

Agent self-improvement need not begin by changing the agent. When failures come from misunderstanding the world, the missing component is an environment belief that evidence can challenge. EvoBelief keeps that belief as falsifiable hypotheses, promotes them into rules only through independent evidence, and lets counterexamples revoke that authority when the environment drifts. Validated rules guide predictions and plans, while unresolved hypotheses can motivate live experiments. Across the evaluated regimes, the results demonstrate improved task performance, reuse of learned rules, adaptation to hidden dynamics changes, and quantitative law discovery, while keeping the belief state open to inspection. Where episodic memory already supplies what the prior lacks, rules and memory act partly as substitutes. Because a validated rule describes the environment rather than a task, one verified correction can improve many tasks at once, including tasks the agent has never seen. Retiring stale rules complements learning new ones, because the ability to withdraw authority keeps accumulated knowledge an asset rather than a liability when the environment changes. More broadly, language models make hypotheses cheap to generate but not to verify, so an agent should act on a belief only after it has survived a test it could have failed. Carrying this discipline to multimodal, embodied, and real-world software agents is a next step.

%% file: sections/limitations.tex
\section{Limitations and Future Work}
\label{app:limitations}

Our experiments use text-based environments, where transitions can be logged and predictions checked when the required outcomes are observable. The belief lifecycle itself is not tied to text, and extending it to multimodal, embodied, and real-world software agents is a natural next step. Such settings bring stochastic effects, which call for statistical tests in place of count thresholds when committing a rule. They also involve interventions that can be costly or irreversible, where live experiments would run under explicit risk limits or human approval. When episodic memory already supplies what the prior lacks, it overlaps with explicit beliefs, so learning when to rely on each is a promising direction. Finally, because validated beliefs carry their evidence records, they could be shared across agents or distilled into model weights, turning what one agent has verified into the prior of the next.

%% file: sections/appendix.tex
\section{Method Details}
\label{app:method-details}

This appendix collects implementation-level mechanisms that \S\ref{sec:object}--\S\ref{sec:testing} reference but do not require for the main argument.

\paragraph{Hypothesis identity and the evidence ledger.}
Hypothesis identity is goal-free: a canonical signature over $(C,A,Y)$ merges rediscoveries of the same claim across different tasks and evidence polarities, so evidence gathered under one goal accrues to the same hypothesis when it resurfaces under another. The per-hypothesis evidence record of \S\ref{sec:object} is implemented as an append-only \emph{evidence ledger}; entries are never rewritten, each entry retains the episode, binding, and test it came from, and every lifecycle transition cites the ledger entries that caused it. Hypotheses cover action-conditioned transitions, object properties and measurement procedures, multi-step procedures, and conditional constraints, the negative counterpart that says what an action \emph{cannot} do while its condition is witnessed.

\paragraph{Evidence admissibility.}
Transitions whose outcome reflects the harness rather than the environment, such as parser rejections, interface errors, and transport failures, are excluded from hypothesis formation and from settlement. A prediction mismatch is admissible only when it is grounded in the realized environment observation itself, not in the agent's own summary of it.

\paragraph{Discriminating tests and counting.}
Rival hypotheses are settled against the same realized evidence, each against the prediction it froze before the test ran, so one measurement can support one rival and counterexample the others. An outcome that every rival predicts discriminates nothing and supports no conditional claim. All readouts of one experimental design count as a single independent test, so a multi-step experiment cannot inflate the support counts of the commitment gate in \S\ref{sec:lifecycle}. Per-episode budgets bound hypothesis formation, the number of experiments, and the steps they may spend, and a hypothesis formed in episode $k$ can be validated only by an independent later occurrence.

\paragraph{Experiment admission.}
Beyond the admission conditions of \S\ref{sec:testing}, an experiment is admitted only when at least two plausible outcomes would lead to different next actions and no public readout already answers the question. Admitted experiments are executed in the order in which their questions become decision-relevant, and an experiment whose question is resolved by other evidence is closed without running.

\paragraph{Commitment thresholds.}
Every instantiation fixes the thresholds of Eq.~\ref{eq:commit} before evaluation and never tunes them on a target. Reuse across episodes requires supports on at least two distinct bindings in at least two distinct episodes, at least one of which is independent of any decision the hypothesis itself influenced. Where a task instance is a single long episode, as in DiscoveryWorld, a hypothesis is committed by one pre-registered held-out test whose prediction was frozen before the test ran.

\section{Language-Model Calls, Prompts, and Retrieval}
\label{app:prompts}

\paragraph{Division of labor.} The framework in Section~\ref{sec:method} specifies a shared belief lifecycle, and the four benchmarks instantiate that lifecycle through different structured interfaces. The frozen language model proposes task actions, supplies anchor predictions, and, where needed, extracts typed observations. The benchmark adapter turns public transitions or measurements into the hypothesis representation used by $W_t$, compiles admissible tests, and enforces evidence accounting, commitment, applicability, retirement, and rule-based planning. Thus the rule lifecycle and its authority checks are implemented outside the actor rather than delegated to a prompt. On ALFWorld, MiniGrid, and ScienceWorld, rule-supported consequences are computed directly from the structured representation. DiscoveryWorld is the only benchmark whose frozen predictor engine may also make a rule-conditioned model call. The actor itself never receives the validated rule text. Table~\ref{tab:instantiations} summarizes the benchmark-specific instantiations, and the remainder of this appendix lists the model calls and prompts. This appendix reproduces the actor, candidate-generation, predictor, and transition-analysis prompts, quoted verbatim from the frozen code of the sealed runs, and describes the DiscoveryWorld knowledge judge. User messages are described by their input fields. Auxiliary engine calls are included in the call inventory.

In DiscoveryWorld's frozen predictor engine, the rule-conditioned prediction is
\begin{equation}
    p_t^W(a)=F_\theta(a,g,z_t,R^{\mathrm{E}}_t(a)),
    \label{eq:pw}
\end{equation}
where $R^{\mathrm{E}}_t(a)$ contains the engine's own applicable rules, distinct from the validated rules $R_t(a)$ of Algorithm~\ref{alg:evobelief}; only rule-supported changes are retained. This call is separate from the law layer's direct consequence prediction and law-derived execution plans.

\begin{table}[t]
\centering\small
\caption{Hypothesis source, behavioral route, experiment route, and use of the frozen engine's rule-conditioned model call on each benchmark. Calls per decision by role are in Table~\ref{tab:llm-calls}.}
\label{tab:instantiations}
\setlength{\tabcolsep}{4pt}
\begin{tabular}{@{}l>{\raggedright\arraybackslash}p{2.7cm}>{\raggedright\arraybackslash}p{3.4cm}>{\raggedright\arraybackslash}p{2.6cm}>{\raggedright\arraybackslash}p{1.7cm}@{}}
\toprule
Benchmark & Hypothesis source & Behavioral route & Experiment route & Rule-conditioned call \\
\midrule
ALFWorld & Typed transitions; code induction & Rule-based planning; precondition veto & Passive evidence only & None \\
MiniGrid & Typed transitions; code induction & Validated plan steps replace the draft & Live experiments scheduled by code & None \\
ScienceWorld & Public observations and typed transitions; code induction & Rule-based control, observation obligations, and veto & Live experiments admitted by code & None \\
DiscoveryWorld & Laws fitted by code from measurements and observations & Code-projected consequences; law-derived execution plans & Compiled live experiments & Conditional \\
\bottomrule
\end{tabular}
\end{table}

\subsection{Roles and Calls}
\label{app:call-inventory}

Table~\ref{tab:llm-calls} lists the calls per decision by role. On every benchmark, the actor prompt contains no rule, prediction, or plan; it receives the goal, the observation, the action interface, its own history, and, in the transition-memory setting, the shared memory block of Section~\ref{app:actor-memory}. Where a rule-conditioned model call is used, it uses the same predictor prompt as the anchor call, with the rule field populated:
\[
p_t^0(a)=\text{predictor, rules omitted},\quad p_t^W(a)=\text{predictor, applicable rules supplied}.
\]
The model's hypothesis-proposal, distillation, and binding calls feed the frozen predictor engine's own rule store, which is consumed only on DiscoveryWorld.

\begin{table}[h]
\centering
\footnotesize
\caption{\textbf{Language-model calls per decision, by role.} $N$ is the number of W-blind alternative candidates (at most 4 on ScienceWorld, 2 on DiscoveryWorld). Per-episode auxiliary calls are a state seed at episode start, hypothesis proposal after a mismatch, trajectory distillation after a success, and role binding for the engine's own experiments. Call counts are taken from one configuration per benchmark, Haiku 4.5 on ALFWorld and MiniGrid and GPT-5.4 on ScienceWorld and DiscoveryWorld; the roles and calls per decision do not depend on the backbone.}
\label{tab:llm-calls}
\setlength{\tabcolsep}{5pt}
\begin{tabular}{@{}lccccc@{}}
\toprule
Benchmark & Actor & Alternatives & Anchor & Rule-conditioned & Analysis \\
\midrule
ALFWorld & 1 & code & 1 (batched) & none & code \\
MiniGrid & 1 & code & 1 per candidate (5) & none & 1 \\
ScienceWorld & 1 & 1 & $1+N$ & none & 1 \\
DiscoveryWorld & 1 & 1 & $1+N$ & up to $1+N$ & 1 \\
\bottomrule
\end{tabular}
\end{table}

\paragraph{Benchmark adapters.} ALFWorld makes exactly two calls per decision, the actor draft and one batched anchor-prediction call over the whole candidate slate (6,172 logical calls over 3,086 environment steps in the sealed Haiku run with transition memory, a ratio of 2.000 in every episode); its candidates are instantiated from the public action grammar by code, and its typed public state is projected by code rather than by a model. MiniGrid issues one anchor-prediction call per candidate and one evidence-analysis call per transition, about 8.5 calls per decision in a sealed episode. ScienceWorld adds one alternatives call, about six calls in a routine decision. DiscoveryWorld uses the benchmark's own ReAct actor prompt unchanged and averages 5.02 logical calls per executed environment action; it is the one benchmark on which the frozen engine's rule-conditioned prediction is requested, and the call is skipped for a candidate when no retrieved rule applies to it. The Knowledge score on DiscoveryWorld is computed after the episode by the benchmark's judge prompt, which never feeds back into the agent. Decoding settings are matched within each benchmark, except for the Hypothesizer's higher output limit: temperature 0 everywhere, with per-role output ceilings fixed in the frozen configuration (actor 256 tokens on ALFWorld, 768 on MiniGrid, 160 on ScienceWorld, and 3,072 on DiscoveryWorld; predictor 768, or 2,048 for the batched ALFWorld call; evidence analysis 768 on MiniGrid, 1,536 on ScienceWorld, and 8,192 on DiscoveryWorld; hypothesis proposal 1,536) and a 60,000-character input ceiling per role.

\subsection{Canonical Prompts by Role}
\label{app:prompt-templates}
\definecolor{evoblue}{RGB}{46,95,158}
\definecolor{evobluebg}{RGB}{244,248,252}
\newtcolorbox{promptbox}[1]{enhanced, breakable, colback=evobluebg, colframe=evoblue, colbacktitle=evoblue, coltitle=white, fonttitle=\bfseries\small, title={#1}, arc=1.5mm, boxrule=0.6pt, left=6pt, right=6pt, top=4pt, bottom=4pt, toptitle=2pt, bottomtitle=2pt, fontupper=\small\raggedright, before skip=4pt, after skip=6pt}

\paragraph{Actor.} The actor chooses one executable action toward the goal from observable information only.
\begin{promptbox}{Actor prompt}
You are an embodied task agent. Use only the public goal, current public observation, realized history, and any explicitly labeled method-visible memory. Never assume hidden evaluator state. Return strict JSON.
\end{promptbox}
Input, in order: the action interface (the grammar or the currently available actions), the optional memory block, the goal, the current observation, and the last 24 rows of the agent's own history as \texttt{Observation:} and \texttt{Action:} lines. Output: one JSON object \texttt{\{"action": "..."\}}. On ScienceWorld the same instruction is issued in ReAct form (``Use only observable information. Never assume hidden state or use reward/scorer internals.'') and answered in two lines, \texttt{Thought:} and \texttt{Action:}, with a 160-token budget; on DiscoveryWorld the benchmark's distributed ReAct prompt is used unchanged, pinned by its digest. When the EvoBelief layer replaces the draft, the executed action enters this history as an ordinary \texttt{Action:} row with its realized observation; the actor is never told that its draft was overridden.

\paragraph{Candidate alternatives.} Where alternatives are proposed by the model (ScienceWorld and DiscoveryWorld), the call is W-blind by construction.
\begin{promptbox}{Candidate-alternatives prompt}
You generate a small finite set of alternative task actions. You have no world-model rules and must not assume hidden state. Return exactly JSON \texttt{\{candidates:[\{action:"...",\allowbreak reason:"..."\}]\}}. Use only the supplied goal, current observation, grounded state claims, recent executed history, \texttt{baseline\_action}, and action grammar. Propose at most the supplied limit, only actions distinct from \texttt{baseline\_action}. Maximize useful candidate coverage: include distinct grounded action templates when appropriate, and for a multi-argument grammar action consider distinct grounded argument assignments or orientations instead of paraphrasing one choice. Prioritize an explicit missing goal prerequisite, recovery from a blocked/repeated action, or an evidence-grounded terminal action. Copy every concrete referent from the supplied surfaces. Do not invent an entity, action, observation, scientific fact, rule, or valid-action oracle. An empty list is valid.
\end{promptbox}
Input: \texttt{goal}, \texttt{observation}, \texttt{state\_claims}, the last six executed steps, \texttt{baseline\_action} (the actor's draft), \texttt{action\_grammar}, \texttt{limit}. On ALFWorld and MiniGrid the alternatives are instantiated from the public action grammar over already-observed entities by code.

\paragraph{Predictor.} The predictor evaluates exactly the supplied action and never proposes or selects one.
\begin{promptbox}{Predictor prompt}
Evaluate exactly the supplied \texttt{draft\_action}; do not propose, edit, select, authorize, or recommend an action. Learned world-model rules may affect prediction only. Return exactly JSON \texttt{\{expected\_outcome:"...",\allowbreak prediction\_rule\_ids:[...],\allowbreak action\_rule\_ids:[],\allowbreak critical\_unknown:null|"...",\allowbreak predicted\_observable\_change:null|"...",\allowbreak action\_feasibility:"feasible|blocked|unknown",\allowbreak goal\_alignment:"advances|neutral|conflicts|unknown"\}}. \texttt{prediction\_rule\_ids} may cite only supplied prediction-authority rules that materially support this prediction. \texttt{action\_rule\_ids} MUST always be empty: W has no action-proposal or action-authority surface. Use only observable state and supplied rules; never use hidden state, score, reward, or a valid-action oracle.
\end{promptbox}
Input: \texttt{goal}, \texttt{observation}, \texttt{state} (the typed claims of $z_t$), \texttt{unknowns}, \texttt{draft\_action}, and \texttt{validated\_rules}, which is empty for $p_t^0(a)$ and holds the rules that passed the applicability check for $p_t^W(a)$. Output contract: expected outcome, observable change, feasibility, goal effect, a critical unknown if any, and the rules cited. On ALFWorld the same contract is requested once for the whole candidate slate, with one entry per candidate and the observable change expressed as a typed delta over the benchmark's public schema. 

\paragraph{Transition and evidence analysis.} After every executed transition the model updates the typed state estimate $z_t$ from observation spans and reports whether the anchor prediction matched. Its verdict is recorded; whether it counts as evidence for any rule is decided by code.
\begin{promptbox}{Transition and evidence analysis prompt}
After a realized transition, use only the supplied \texttt{observation\_spans} and return exactly JSON with \texttt{\{state\_update:\{added\_claims:[\{text,\allowbreak span\_id\}],\allowbreak superseded\_claims:[\{new\_text,\allowbreak span\_id,\allowbreak supersedes:[claim\_id]\}],\allowbreak maintenance\_pruned\_claims:[claim\_id]\},\allowbreak prediction\_verdict:"match|mismatch|unscorable",\allowbreak evidence\_span\_id:null|"OBS\_NNN"\}}. Every non-null span ID must be copied from \texttt{observation\_spans}. Do not retype quotes. Validate each state-update list independently: when no row is grounded, return an empty list. Never infer hidden state, reward, score, completion, action validity, or unobserved effects.
\end{promptbox}
Input: the prior claims, the observation before, the executed action, the observation after, the prediction, and the observation spans with identifiers. A repair variant re-requests only the JSON framing of a malformed reply. At episode start a seed call of the same form returns the initial claims and unknowns (at most 32 and 12) from the first observation. On ALFWorld the typed public state is projected by code and this role is not called.

\paragraph{Knowledge judge (DiscoveryWorld).} The benchmark's own scoring prompt, used unchanged after the episode: it receives the task description, the agent's knowledge, and one critical question, and answers \texttt{1} or \texttt{0} with an explanation; the benchmark asks two critical questions per task instance, one call each. For EvoBelief the agent's knowledge is the deterministic rendering of the store, in which a statement is listed as established only if its prediction was frozen before the test, the realized evidence settled it as passed, and no counterexample stands. The rendered knowledge never names an arm, and every prompt is checked for that before dispatch.

\subsection{Actor-Side Transition Memory}
\label{app:actor-memory}

The setting ``with transition memory'' installs one shared factor on the actor of every method. Before the actor is asked for an action, the factor retrieves at most four realized public transitions from earlier episodes, ranked by lexical overlap between the current goal and public state and the stored goal and pre-state, with a per-item limit of 1,200 characters and a total budget of 4,800. Only transitions of completed earlier episodes are retrievable, every realized transition is stored regardless of its outcome, no outcome label enters the store, and the candidate action never enters the query. Each item is rendered as its goal, the observation before, the action, the observation after, and the public delta, under the header \texttt{Method-visible memory:} (\texttt{Cross-episode public transition memory (previous episodes only):} on ALFWorld), spliced into the actor prompt between the action interface and the history. The factor reads no rule, no store, and no arm name, and a run refuses to start if its budget differs from the frozen one. On MiniGrid the EvoBelief runs use the setting without memory.

\section{Decision-Step Algorithm}
\label{app:algorithm}

Algorithm~\ref{alg:evobelief} summarizes one decision step of \S\ref{sec:consumption}. Candidates are formed without access to $W_t$; each receives an anchor prediction and an effective prediction $\hat p_t(a)$ that retains only rule-supported changes to $p_t^0(a)$; where a rule-conditioned model output (Eq.~\ref{eq:pw}) is available, the changes it proposes are retained only where a validated rule supports them. An experiment may be admitted at the current step when a decision-relevant claim is unresolved, its setup is executable, and budget remains (\S\ref{sec:testing}); admission freezes the test criteria and does not by itself give the experiment priority. \textsc{Select} follows the fixed admission, retention, and priority rules of the instantiation: it returns a candidate action, the first step of a validated plan, or the next step of an admitted experiment, together with the basis $d_t$ for that choice, and it preserves the $W$-blind choice when no permitted intervention applies. \textsc{Plan} checks rule conditions and scope at every predicted intermediate state, and its first step need not lie in $A_t$. The decision record $d_t$ is formed before execution and preserves the predictions and evidence provenance needed to distinguish independent tests from rule-caused confirmations. After the environment step, \textsc{Absorb} updates the experiment's progress from the executed transition, settles tests only when their outcome can be assessed, and recomputes the lifecycle of \S\ref{sec:lifecycle}; unfinished tests remain pending. Every route spends the same task budget. The implementations of \textsc{Consequence}, \textsc{Plan}, and \textsc{Select}, including the use of rule-conditioned model calls, are specified in \S\ref{sec:setup} and Appendix~\ref{app:prompts}.

\begin{algorithm}[h]
\caption{One EvoBelief decision step}
\label{alg:evobelief}
\begin{algorithmic}[1]
\Require goal $g$, frozen model $F_\theta$, history $h_t$, state estimate $z_t$, world belief $W_t$, pending experiment $E_t$ (possibly empty), remaining environment-action budget $B_t>0$
\State $A_t\gets\Call{Candidates}{g,h_t,z_t}$ \Comment{formed without access to $W_t$}
\ForAll{$a\in A_t$}
    \State $p_t^0(a)\gets F_\theta(a,g,z_t)$ \Comment{anchor prediction}
    \State $R_t(a)\gets\Call{Applicable}{W_t,z_t,a}$ \Comment{validated rules whose condition holds and whose scope covers the binding}
    \State $\hat p_t(a)\gets\Call{Consequence}{z_t,a,g,p_t^0(a),R_t(a)}$ \Comment{rule-supported changes only; $p_t^0(a)$ if $R_t(a)=\varnothing$}
\EndFor
\State $E_t\gets\Call{Admit}{W_t,z_t,g,B_t,E_t}$ \Comment{open or keep an experiment: decision-relevant, executable, within budget; criteria frozen}
\State $\pi_t\gets\Call{Plan}{W_t,z_t,g}$ \Comment{intermediate states predicted by validated rules; empty if no rule applies}
\State $(a_t,d_t)\gets\Call{Select}{A_t,\{p_t^0(a)\},\{\hat p_t(a)\},\pi_t,E_t,z_t,g}$ \Comment{a candidate, the first step of $\pi_t$, or the next step of $E_t$, with its basis}
\State $o_{t+1}\gets\Call{Env}{a_t}$;\quad $B_{t+1}\gets B_t-1$ \Comment{every route spends the task budget}
\State $(z_{t+1},e_t)\gets\Call{Encode}{z_t,a_t,o_{t+1}}$
\State $(W_{t+1},E_{t+1})\gets\Call{Absorb}{W_t,E_t,e_t,d_t}$ \Comment{form; advance or settle tests; commit, narrow, demote, retire}
\State record $d_t$ and the realized outcome $o_{t+1}$
\end{algorithmic}
\end{algorithm}

\section{Experimental Setup Details}
\label{app:setup-details}

\begin{figure}[t]
    \centering
    \includegraphics[width=\textwidth]{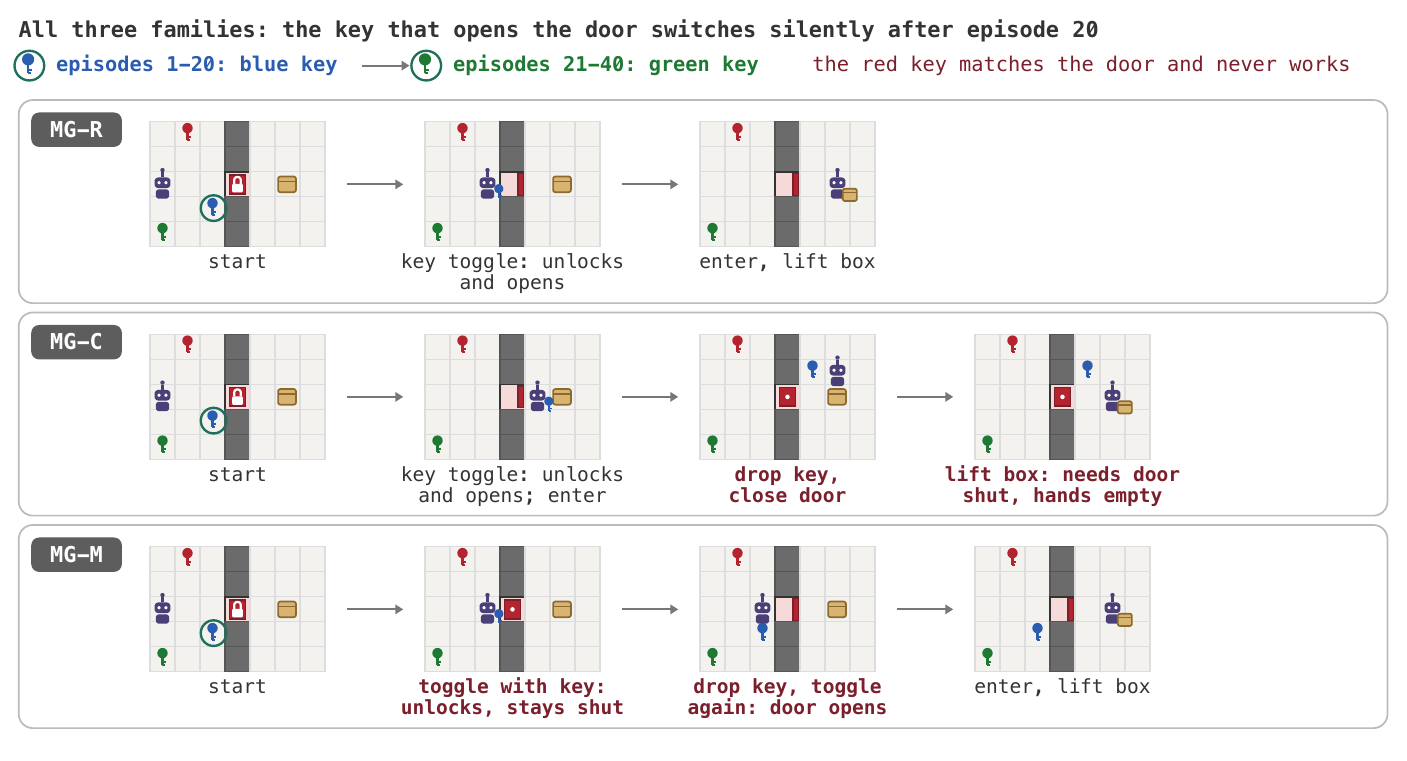}
    \caption{\textbf{What each MiniGrid family requires.} One row per family, each frame a state of the two-room grid: the robot is the agent, a key drawn next to it is the key it carries, the box drawn next to it is the lifted target. A padlock marks a locked door, a plain door is shut but unlocked, and a door drawn aside is open. In every family the red key matches the door and never works, the blue key opens the door for the first twenty episodes, and after episode 20 the green key does instead; the ring marks the key that opens the door and the switch is never announced. In MG-R (R for rebind) one toggle with the working key unlocks and opens the door, and nothing more is asked. In MG-C (C for conjunctive) the door behaves the same, but the box lifts only with the door shut and the agent's hands empty, so after entering the agent must drop the key and close the door it just opened. In MG-M (M for multi-stage) toggling with the working key only unlocks the door, which stays shut until a second toggle with empty hands. The switch is the only change within a stream; the family rules hold in both halves. The steps in dark red are the ones the pretrained prior does not expect.}
    \label{fig:minigrid-storyboard}
\end{figure}

\paragraph{Benchmarks.} The static, prior-aligned regime uses \textbf{ScienceWorld} \citep{wang2022scienceworld}, a text-based science laboratory whose mechanics are stationary but partially hidden (90 episodes per method), and \textbf{ALFWorld} \citep{shridhar2021alfworld}, a household suite testing beliefs about object state, affordances, and procedures (the 134 unseen test tasks, each run once per method). The dynamic regime uses \textbf{MiniGrid} \citep{chevalier2023minigrid}, where the agent must identify the key that actually opens a locked door, cross into the far room, and lift a target box within 24 steps (Figures~\ref{fig:minigrid-env} and~\ref{fig:minigrid-storyboard} describe the families). A seed fixes the sequence of layouts of a stream: key, agent, and box locations are re-sampled every episode, with the first three episodes on the three fixed layouts of the original benchmark. The campaign runs MG-C and MG-M on three development seeds, on three held-out seeds fixed before the runs, on the same held-out seeds with every color and room name permuted in the observations, and on a third seed set reserved for the mechanism ablations, giving 240 matched episodes per method per set; MG-R runs on three further fresh seeds, 120 matched episodes per method. The discovery regime uses \textbf{DiscoveryWorld} \citep{jansen2024discoveryworld} on two themes. In Reactor Lab, each reactor activates only when tuned to within $\pm 2$\,Hz of a resonance frequency governed by a hidden parametric law of a crystal property measurable through instruments, so the held-out resonance of a new reactor cannot be retrieved from memory and must be computed from a fitted law. In Plant Nutrients, a field's soil controller sets nutrient levels, a planted seed grows only under the hidden nutrient rule, and the Challenge tier hides that rule in a logical combination of two nutrients. Both themes run at their Normal and Challenge tiers on GPT-5.4: twelve fresh task instances per tier, frozen before any run, four methods on the same task instances, one attempt each (provider-side failures were re-run under the pre-declared transport policy), 320 environment actions and 1,650 model calls per task instance, and no transition memory. On Reactor Lab, EvoBelief's experiments draw on a science budget of 208 of the 320 actions. Completion and the subgoal-based procedural score are the benchmark's own, and Knowledge is the benchmark's LLM-judged score over two sealed critical questions, judged blind to the method with one judge configuration shared by both themes. The benchmark's own Hypothesizer agent is included as a baseline; its rows in Table~\ref{tab:discoveryworld-results} come from a separately frozen re-run whose only change is a raised output ceiling (Appendix~\ref{app:dw-details}). A frozen-belief diagnostic and the Plant Nutrients protocol and mechanism are reported in Appendix~\ref{app:dw-details}.

\paragraph{Actor settings.} ALFWorld and ScienceWorld run under two actor settings, without and with \emph{transition memory}; on MiniGrid the EvoBelief runs use the setting without it; no method uses it on DiscoveryWorld. Transition memory is specified in Appendix~\ref{app:actor-memory}.

\paragraph{Baselines.}
\textbf{ReAct} keeps no persistent cross-episode belief \citep{yao2023react}. \textbf{ITP-I} consults imagine-then-plan lookahead with a learned world model \citep{liu2026itp}. \textbf{WorldEvolver} evolves a world model online over stored transitions and semantic memory, a closely related self-evolving world-model baseline \citep{zhang2026worldevolver}. WorldEvolver and ITP-I are the official released implementations, run through the same harness, interface, and budgets as every other method. WALL-E 2.0's published ALFWorld result uses a different rule-learning protocol and backbone and is not directly comparable to our single-pass benchmark evaluation. \textbf{Frozen WM} is the same W-blind actor with the same anchor predictor and selector, but its predictions and choices follow the pretrained prior alone: no learned rule may alter a prediction, a ranking, or a plan. It does not run EvoBelief's belief store. Frozen WM and EvoBelief are independent live runs paired by task, not replays. For rule-guided interventions we record the rule, the prediction or plan it changed, and the realized outcome. The paired comparison with Frozen WM evaluates the complete EvoBelief system, including its experiments, relative to the matched prior-driven control.

\paragraph{Protocol and fairness.}
WorldEvolver's original evaluation reports best-of-$L$ success over $L{=}5$ trials per task, so a task counts as solved when any one of the five trials succeeds \citep{zhang2026worldevolver}. In the main benchmark comparisons, each task is attempted once. All methods, including the baselines we rerun, share one harness with the same task set, task order, seeds, action interface, context limits, retry policy, and scoring. Both backbones run the identical protocol, and success is task completion as scored by the benchmark. On ScienceWorld with GPT-5.4, EvoBelief's run without transition memory was preregistered, executed blind, and sealed first. The Frozen WM and WorldEvolver comparators, and then all three runs with transition memory, were authorized afterwards and executed blind on the same 90 tasks, each group unblinded once.

\paragraph{Metrics and controls.}
MiniGrid is split at the hidden changepoint into pre-drift and post-drift halves. Mechanism diagnostics count rules formed, tested, supported, counterexampled, and retired, and \emph{decision flips}, decisions where the anchor prediction alone would have selected differently and a validated rule's authorized change is the recorded cause. Held-out structured state-change prediction uses an identical frozen prediction interface for every method. The frozen predictor engine that supplies the anchor prediction, shared with the Frozen WM control, also runs sandboxed tests of its own typed rules on ScienceWorld and ALFWorld. These tests do not replace live task actions, and their outcomes do not enter EvoBelief's store.

\section{Additional Results}
\label{app:additional-results}

\subsection{Mechanism Summary}
\label{app:mechanism-summary}

Table~\ref{tab:mechanism-summary} collects, in one vocabulary, what the belief machinery did on each benchmark, and \S\ref{sec:analysis} summarizes how the routes of \S\ref{sec:consumption} carry different loads by regime. The counts are reported in the unit each implementation records, so the columns are comparable in kind rather than in magnitude: an ALFWorld hypothesis is a rule branch over object roles, a MiniGrid hypothesis is a guarded rule instance that is re-validated after every epoch, a ScienceWorld hypothesis is a typed rule, and a DiscoveryWorld hypothesis is a candidate law.

\begin{table}[h]
\centering
\footnotesize
\setlength{\tabcolsep}{3pt}
\caption{\textbf{What the belief machinery did, by benchmark.} One setting per benchmark: ALFWorld, Haiku 4.5 without transition memory (134 tasks); MiniGrid, the six development streams (240 episodes); ScienceWorld, GPT-5.4 without transition memory (90 tasks); DiscoveryWorld Reactor Lab and Plant Nutrients, GPT-5.4 (24 task instances each). Each row counts the unit named in its label, and the rows are not stages of one funnel. ALFWorld: 10 validated rules over 43 validation events. MiniGrid: 220 distinct validated rules over 2,185 validated rule versions, and counterexamples are negative witness records. ScienceWorld: 54 experiments opened, 39 completed, yielding 573 support records, while the 4,031 passive supports come from ordinary task transitions. Supports are counted per occurrence, so a claim observed repeatedly in one context adds many records but little independent evidence, which is why validated rules are far fewer than support records. Confirmed: the predicted effect held in the next observation, over the decisions a rule changed on ALFWorld, all plan steps on MiniGrid, the scorable rule predictions on ScienceWorld, law-derived plans on Reactor Lab, and positive consequence verifications on Plant Nutrients. On DiscoveryWorld, the counterexample row counts rival laws falsified. ScienceWorld's procedural safeguards for task milestones and irreversible steps, fixed before evaluation (\S\ref{sec:setup}), are not belief actions and are not counted. A dash means the implementation does not record the quantity.}
\label{tab:mechanism-summary}
\begin{tabular}{@{}lrrrrr@{}}
\toprule
 & ALFWorld & MiniGrid & ScienceWorld & DW Reactor & DW Plant \\
\midrule
Hypotheses formed (claims) & 12 & 639 & 236 & 669 & 324 \\
Passive supports (records) & 230 & 2,156 & 4,031 & -- & -- \\
Live experiments (runs) & 0 & 458 & 54 & 96 & 51 \\
Validated rules (distinct) & 10 & 220 & 36 & 18 & 38 \\
Counterexamples (records) & 9 & 10,645 & 21 & 296 & 12 \\
Retirements (rule versions) & 2 & 845 & 70 & -- & 15 \\
\midrule
Decisions under validated rules & 707 & 886 & 14 & 209 & 1,129 \\
\quad unit & actions & plan steps & actions & actions & actions \\
Experiment steps (live actions) & 0 & 1,051 & 424 & -- & -- \\
Confirmed & 265/324 & 830/886 & 13/14 & 17/18 & 36 \\
\quad unit & decisions & plan steps & predictions & plans & verifications \\
\bottomrule
\end{tabular}
\end{table}

\subsection{Paired Tests and Confidence Intervals}
\label{app:paired-stats}

Because every comparison is paired, Table~\ref{tab:paired-stats} reports, for each matched comparison, the exact two-sided McNemar test on the discordant pairs and a percentile bootstrap interval for the difference in success rate, resampling paired episodes ten thousand times. The DiscoveryWorld rows are the sealed campaign's own instance-paired analysis.

\begin{table}[t]
\centering
\scriptsize
\setlength{\tabcolsep}{3pt}
\caption{\textbf{Paired tests for the matched comparisons.} Success rates in percent; $\Delta$ is EvoBelief minus the baseline in percentage points with its 95\% percentile bootstrap interval over paired episodes; Ours and Base are the two success rates, Disc. counts the episodes only EvoBelief solved against those only the baseline solved; $p$ is the exact two-sided McNemar test on those pairs. ALFWorld pairs the 134 unseen tasks, WorldEvolver being the authors' release; ScienceWorld pairs its 90 tasks; MiniGrid pairs episodes by stream and position, 240 per seed set, and the post-versus-pre rows compare EvoBelief's second half with its own first half, pairing episode $i+20$ with episode $i$ of the same stream; MiniGrid episodes within a stream share an evolving belief state, and the episode-level tests and intervals do not account for this dependence; DiscoveryWorld pairs the 12 fresh task instances per tier of each theme, with the Hypothesizer at its raised output ceiling.}
\label{tab:paired-stats}
\begin{tabular}{@{}llrrrrrrr@{}}
\toprule
Benchmark, setting & Baseline & $n$ & Ours & Base & $\Delta$ & 95\% CI & Disc. & $p$ \\
\midrule
ALFWorld, Haiku, no memory & Frozen WM & 134 & 66.42 & 42.54 & 23.88 & 16.42, 31.34 & 34 / 2 & $<$0.001 \\
 & WorldEvolver & 134 & 66.42 & 35.07 & 31.34 & 21.64, 41.04 & 51 / 9 & $<$0.001 \\
ALFWorld, Haiku, memory & Frozen WM & 134 & 68.66 & 57.46 & 11.19 & 2.99, 20.15 & 25 / 10 & 0.017 \\
 & WorldEvolver & 134 & 68.66 & 58.96 & 9.70 & 1.49, 17.91 & 24 / 11 & 0.041 \\
ALFWorld, GPT-5.4, no memory & Frozen WM & 134 & 76.12 & 44.78 & 31.34 & 23.13, 40.30 & 45 / 3 & $<$0.001 \\
 & WorldEvolver & 134 & 76.12 & 61.94 & 14.18 & 5.22, 23.13 & 29 / 10 & 0.003 \\
ALFWorld, GPT-5.4, memory & Frozen WM & 134 & 76.87 & 55.22 & 21.64 & 13.43, 29.85 & 33 / 4 & $<$0.001 \\
 & WorldEvolver & 134 & 76.87 & 72.39 & 4.48 & $-$2.99, 12.69 & 18 / 12 & 0.362 \\
\midrule
ScienceWorld, Haiku, no memory & Frozen WM & 90 & 64.44 & 46.67 & 17.78 & 5.56, 30.00 & 25 / 9 & 0.009 \\
 & WorldEvolver & 90 & 64.44 & 52.22 & 12.22 & 1.11, 23.33 & 19 / 8 & 0.052 \\
ScienceWorld, Haiku, memory & Frozen WM & 90 & 68.89 & 47.78 & 21.11 & 7.78, 34.44 & 29 / 10 & 0.003 \\
 & WorldEvolver & 90 & 68.89 & 64.44 & 4.44 & $-$7.78, 16.67 & 18 / 14 & 0.597 \\
ScienceWorld, GPT-5.4, no memory & Frozen WM & 90 & 71.11 & 53.33 & 17.78 & 4.44, 31.11 & 28 / 12 & 0.017 \\
 & WorldEvolver & 90 & 71.11 & 65.56 & 5.56 & $-$4.44, 15.56 & 14 / 9 & 0.405 \\
ScienceWorld, GPT-5.4, memory & Frozen WM & 90 & 72.22 & 58.89 & 13.33 & 3.33, 23.33 & 17 / 5 & 0.017 \\
 & WorldEvolver & 90 & 72.22 & 67.78 & 4.44 & $-$6.67, 15.56 & 16 / 12 & 0.572 \\
\midrule
MiniGrid, development seeds & Frozen WM & 240 & 79.17 & 0.42 & 78.75 & 73.33, 83.75 & 189 / 0 & $<$0.001 \\
 & WorldEvolver & 240 & 79.17 & 5.83 & 73.33 & 67.50, 79.17 & 178 / 2 & $<$0.001 \\
MiniGrid, held-out seeds & Frozen WM & 240 & 69.58 & 1.25 & 68.33 & 62.08, 74.17 & 165 / 1 & $<$0.001 \\
MiniGrid, permuted labels & Frozen WM & 240 & 80.83 & 7.08 & 73.75 & 68.33, 79.17 & 177 / 0 & $<$0.001 \\
MiniGrid, development, post vs pre & pre-switch & 120 & 88.33 & 70.00 & 18.33 & 10.83, 25.83 & 24 / 2 & $<$0.001 \\
MiniGrid, held-out, post vs pre & pre-switch & 120 & 75.83 & 63.33 & 12.50 & 3.33, 22.50 & 25 / 10 & 0.017 \\
MiniGrid, permuted, post vs pre & pre-switch & 120 & 87.50 & 74.17 & 13.33 & 5.83, 21.67 & 21 / 5 & 0.002 \\
\midrule
DiscoveryWorld Reactor, Normal & ReAct & 12 & 83.33 & 0.00 & 83.33 & 58.33, 100.00 & 10 / 0 & 0.002 \\
 & WorldEvolver & 12 & 83.33 & 0.00 & 83.33 & 58.33, 100.00 & 10 / 0 & 0.002 \\
 & Hypothesizer & 12 & 83.33 & 0.00 & 83.33 & 58.33, 100.00 & 10 / 0 & 0.002 \\
DiscoveryWorld Reactor, Challenge & ReAct & 12 & 58.33 & 0.00 & 58.33 & 33.33, 83.33 & 7 / 0 & 0.016 \\
 & WorldEvolver & 12 & 58.33 & 0.00 & 58.33 & 33.33, 83.33 & 7 / 0 & 0.016 \\
 & Hypothesizer & 12 & 58.33 & 0.00 & 58.33 & 33.33, 83.33 & 7 / 0 & 0.016 \\
\midrule
DiscoveryWorld Plant, Normal & ReAct & 12 & 75.00 & 41.67 & 33.33 & 0.00, 66.67 & 5 / 1 & 0.219 \\
 & WorldEvolver & 12 & 75.00 & 16.67 & 58.33 & 16.67, 91.67 & 8 / 1 & 0.039 \\
 & Hypothesizer & 12 & 75.00 & 50.00 & 25.00 & $-$16.67, 66.67 & 5 / 2 & 0.453 \\
DiscoveryWorld Plant, Challenge & ReAct & 12 & 83.33 & 50.00 & 33.33 & 8.33, 58.33 & 4 / 0 & 0.125 \\
 & WorldEvolver & 12 & 83.33 & 75.00 & 8.33 & $-$16.67, 33.33 & 2 / 1 & 1.000 \\
 & Hypothesizer & 12 & 83.33 & 75.00 & 8.33 & $-$16.67, 33.33 & 2 / 1 & 1.000 \\
\bottomrule
\end{tabular}
\end{table}

\subsection{Regime I: Static-Environment Details}

\paragraph{Regime I details.}
On ALFWorld the gain concentrates on the task families whose procedures the prior misstates, above all cooling, heating, and two-object placement, where a wrong belief about what an action does, or about the order in which two actions must be taken, cannot be repaired by recall alone. Over the deployment (Figure~\ref{fig:alfworld-deployment}), the curves begin to separate after the first rules are committed. The generic household dynamics, that reaching a receptacle reveals its contents, that taking requires an empty hand, that placing moves an object, are committed within the first task type and consumed in every later one, so each new type begins with most of its dynamics already validated; the one process rule a type adds, cooling, heating, or cleaning, is committed within its first few tasks and reused on the rest. The cumulative gap therefore widens with deployment, fastest on the process types where the prior misstates the procedure. A validated rule removes the exploratory and repeated actions the prior spends on rediscovering a precondition it has already met, so EvoBelief solves more tasks with fewer decisions and fewer model calls, whereas WorldEvolver pays for its world model with far more calls than the agents that consult none. Because the benchmark enumerates its task types in blocks, later episodes are also harder for the prior: over the second half of the deployment, where the enumeration reaches the process types, Frozen WM solves fewer tasks than in the first half while EvoBelief solves more. This is why the comparison is drawn on matched tasks: the widening is measured against the same agent on the same task, not against an easier past. Of the 1,115 decisions a validated rule changed across the four settings, 907 had their predicted effect fully confirmed in the next observation. The store classified 199 outcomes as inconclusive: the go-to rule predicted arrival at a receptacle and a description of its contents, the agent arrived, and the receptacle turned out to be closed, so the second atom was unobservable. In the remaining 9, the environment refused the action, and the store recorded the refusal as evidence against the rule's condition or as an inconclusive refusal. None of the 1,115 was classified as an effect mismatch.

\paragraph{ALFWorld: rules travel across tasks.}
The belief persists across the 134 unseen tasks, so a rule validated early is available to every later task. Table~\ref{tab:alfworld-transfer} counts, for every setting, where decisions under a validated rule were made relative to the task in which the rule earned its authority, and Table~\ref{tab:alfworld-transfer-rules} lists the rules of one setting. Uses in the task where the rule was validated are rare by construction, since validation needs independent evidence from at least two tasks.

\begin{table}[t]
\centering
\footnotesize
\setlength{\tabcolsep}{3pt}
\caption{\textbf{Where committed rules are used on ALFWorld.} Rule-use decisions: decisions whose executed action a validated rule licensed. Later task: a task after the one in which the rule was first validated. Other family: the task family differs from that of the task in which the rule was validated. Type unseen at formation: the executed action binds at least one role to an object type absent from every binding that formed the rule (instance numbers ignored). Changed: decisions under a validated rule whose executed action differs from the controller's draft after normalizing the surface forms of put. Confirmed: the transition was not rejected and every atom the rule predicted appears in the next observation. Percentages are of decisions under a validated rule, and of changed decisions for the last column.}
\label{tab:alfworld-transfer}
\begin{tabular}{@{}lrrrrrrr@{}}
\toprule
Setting & Rules & \begin{tabular}[c]{@{}c@{}}Rule-use\\decisions\end{tabular} & Later task & Other family & \begin{tabular}[c]{@{}c@{}}Type unseen\\at formation\end{tabular} & Changed & Confirmed \\
\midrule
Haiku, no memory & 10 & 707 & 99.86 & 82.46 & 88.12 & 324 & 81.79 \\
Haiku, memory & 8 & 753 & 99.87 & 85.79 & 90.17 & 250 & 78.40 \\
GPT, no memory & 8 & 774 & 99.87 & 85.66 & 91.34 & 304 & 83.22 \\
GPT, memory & 8 & 789 & 99.75 & 86.69 & 90.87 & 237 & 81.43 \\
\bottomrule
\end{tabular}
\end{table}

\begin{table}[t]
\centering
\footnotesize
\setlength{\tabcolsep}{3pt}
\caption{\textbf{The rules of one setting, GPT-5.4 without transition memory.} Validation task: the index of the task in the deployment order at which the rule first earned authority, with that task's family. Tasks used: distinct later tasks with a decision taken under the rule. Bindings: distinct object and receptacle instances the rule was bound to. The two go-to rules differ in their effect: one predicts that reaching a receptacle reveals its contents, the other only the new location. Clean, heat, and cool rules can only be used in their own family.}
\label{tab:alfworld-transfer-rules}
\begin{tabular}{@{}llrrrrr@{}}
\toprule
Rule & Validation task & Tasks used & Families & Bindings & Changed & Confirmed \\
\midrule
go to (contents) & 1, look at in light & 115 & 6 & 26 & 221 & 170 \\
take & 2, look at in light & 109 & 6 & 83 & 20 & 20 \\
put & 7, look at in light & 91 & 5 & 63 & 4 & 4 \\
open & 24, pick and place & 63 & 5 & 8 & 8 & 8 \\
go to (location) & 21, pick and place & 33 & 5 & 10 & 33 & 33 \\
clean & 44, clean then place & 25 & 1 & 16 & 4 & 4 \\
heat & 95, heat then place & 18 & 1 & 9 & 11 & 11 \\
cool & 76, cool then place & 13 & 1 & 9 & 3 & 3 \\
\bottomrule
\end{tabular}
\end{table}

\subsection{Regime II: Drift Details}
\label{app:drift-details}

\paragraph{Regime II details.}
The two families MG-C and MG-M were designed so that the solution requires one step the pretrained prior does not expect, and the 24-step budget leaves no room to discover it by chance. The same prior-governed agent is competent where the environment mostly agrees with its prior and helpless where one hidden rule contradicts it, so the near-zero rows of MG-C and MG-M in Table~\ref{tab:minigrid-drift} measure the hidden rules, not a weak baseline. In the first episode of the development stream on MG-C (seed 256), ReAct picks up the color-matched red key, toggles the door three times without effect, cycles through the other keys, opens the door with the blue key on its fourteenth step, enters the far room, and then fails three lifts in a row because the door it opened is still open; it never closes it. On MG-M the same agent unlocks the door on its fifteenth step, sees it closed but unlocked, and spends the remaining steps trying to walk through it. The frozen predictor does not help here: it predicts with the same prior the actor drafts from, so it expects the decoy key to work just as the actor does and never overrules the draft, and all prior-governed methods share a single success. MG-R is the control for this design: it keeps the decoy key and the silent switch, so the prior stops at about half of the episodes, whereas EvoBelief reaches a similar level on all three families, so the learned rules erase a difficulty gap that the prior cannot. On MG-R EvoBelief loses nothing before the switch, the two agents are level only in the first episodes after it, and EvoBelief pulls furthest ahead in the late post-switch episodes once the successor rule is committed, so the gain is not an artifact of a floor. On the development streams of Figure~\ref{fig:minigrid-deployment}, the last dozen episodes of every stream are at or above their pre-switch level, and Frozen WM uses the full budget of 24 decisions in every episode and makes 178 model calls per episode against 98 for EvoBelief. Every method runs under the same harness, budget, seeds, and step limit. Development seeds were used while building the generic instantiation; the held-out seeds, their label permutation, the mechanism seeds, and the MG-R seeds were fixed before those runs.

\begin{table}[t]
\centering
\small
\setlength{\tabcolsep}{8pt}
\caption{\textbf{MiniGrid on held-out seeds (\%).} Haiku 4.5, actor without transition memory. Three seeds on both families, fixed before any run and never used while building the method, 240 episodes per method per block; the permuted block renames every color and room in the observations of both methods. Higher is better.}
\label{tab:minigrid-heldout}
\begin{tabular}{@{}llrrr@{}}
\toprule
Seeds & Method & Pre-switch & Post-switch & All \\
\midrule
Held-out & Frozen WM & 2.50 & 0.00 & 1.25 \\
& \textbf{EvoBelief} & \textbf{63.33} & \textbf{75.83} & \textbf{69.58} \\
\midrule
Held-out, permuted labels & Frozen WM & 10.00 & 4.17 & 7.08 \\
& \textbf{EvoBelief} & \textbf{74.17} & \textbf{87.50} & \textbf{80.83} \\
\bottomrule
\end{tabular}
\end{table}

\begin{table}[t]
\centering
\small
\setlength{\tabcolsep}{8pt}
\caption{\textbf{What the MiniGrid gain needs (\%).} Each block compares variants of EvoBelief on one seed set, 240 episodes per row. The planner without learned rules never leaves the start state of its search and defers to the prior on every decision; without active experiments, rules never gather independent support; in the shadow variant rules are learned and validated but no plan step is executed; without the commitment gate a rule may act on its first supported instance. Higher is better.}
\label{tab:minigrid-ablation}
\begin{tabular}{@{}llrrr@{}}
\toprule
Seeds & Variant & Pre-switch & Post-switch & All \\
\midrule
Development & Planner without learned rules & 0.83 & 0.00 & 0.42 \\
& No active experiments & 0.83 & 0.00 & 0.42 \\
& \textbf{EvoBelief} & \textbf{70.00} & \textbf{88.33} & \textbf{79.17} \\
\midrule
Mechanism & Rules never executed (shadow) & 50.00 & 64.17 & 57.08 \\
& No commitment gate & 66.67 & 70.00 & 68.33 \\
& \textbf{EvoBelief} & \textbf{61.67} & \textbf{77.50} & \textbf{69.58} \\
\bottomrule
\end{tabular}
\end{table}

\paragraph{Experiments double as exploration on MiniGrid.}
In the shadow variant of Table~\ref{tab:minigrid-ablation} the belief forms, tests, and validates rules exactly as in the full method and the planner computes the same plans, but no plan step is executed: every executed action comes from the prior or from an experiment. In episode 7 of MG-C seed 797 twelve of its fourteen decisions are experiment steps, which pick up the working key, open the door, close it again from the far room, and leave the agent one drop away from the box; the prior's draft finishes the task. In episode 16 the full method spends three experiment steps and then executes a four-step committed plan, entering the far room, closing the door, dropping the key, and lifting the box, while the shadow variant needs twelve decisions, eleven of them experiment steps, and succeeds only because the prior's last draft coincides with the plan step it was not allowed to take. The value of executing committed rules is therefore concentrated on MG-C, whose solution is a chain no single experiment completes, and it also shortens episodes and lowers cost. Acting on rules at their first supported instance, without the independent confirmation the commitment gate demands, leaves success and the rate of failed authorized steps close to the full method's on this benchmark, whose public transitions are deterministic; we report it as a null result rather than as support for the gate.

\paragraph{One rule's life, across the silent changepoint.}
The stream on MG-C seed 749 shows the lifecycle doing what drift demands. In episode 5 an active experiment validates a first door rule, that toggling opens the door while the blue key is held; the same episode refutes it, because the rule ignored the door's own state, and it is retired within three decisions. In episode 6 an experiment validates its successor, which conditions on the door being locked and on where the other keys lie, and that rule carries every plan through episode 20. The switch makes it fail at the first decision of episodes 21 and 22; two refutations in distinct episodes open a new evidence epoch for the claim, releasing the pre-switch evidence, and success resumes in episode 24 on experiment and prior steps. In episode 26 an active experiment validates the post-switch successor, that toggling opens the door while the green key is held, which then carries the remaining plans. No grant is ever archival: authority is always backed by currently surviving evidence, so the first generation does not survive its own counterexamples and the successor earns commitment with freshly built supports. On MG-M the same machinery commits the two-stage door as two rules, that the working key unlocks the door and that an empty-handed toggle then opens it, and relearns the first of them after the switch.

\subsection{Regime III: Discovery Details}
\label{app:dw-details}

Table~\ref{tab:dw-mechanism} counts what happened on the twenty-four fresh Reactor Lab task instances, and Table~\ref{tab:dw-laws} lists the validated law of every task instance. A calibration law learned on disjoint instances did not solve the evaluated instances without local experimentation. A frozen belief preloaded with such a law and run without experiments on four Normal and six Challenge task instances names the variable and earns half of the knowledge credit but never the coefficients, and completes nothing.

\begin{table}[t]
\centering
\small
\caption{\textbf{What EvoBelief did on the fresh task instances.} Counts from the sealed science records of the twelve task instances per tier; the baselines have no such record by construction. A hypothesis is a candidate law with a prediction frozen before any experiment. One executed experiment settles every rival whose frozen prediction its realized outcome contradicts, so fewer experiments falsify more rivals, and a law is validated only by a held-out prediction the reactor confirms, never by fit quality alone. Model calls and environment actions are totals over the tier.}
\label{tab:dw-mechanism}
\begin{tabular}{@{}lrr@{}}
\toprule
 & Normal & Challenge \\
\midrule
Hypotheses frozen & 236 & 433 \\
Experiments executed & 48 & 48 \\
Rivals falsified & 157 & 139 \\
Laws validated & 10 & 8 \\
Task instances with a validated law & 10 & 8 \\
Task instances whose law-derived plan passed & 10 & 7 \\
Task instances completed & 10 & 7 \\
Actions taken under a validated law & 115 & 94 \\
\midrule
Model calls, EvoBelief & 2744 & 4988 \\
Model calls, ReAct & 19559 & 18790 \\
Model calls, WorldEvolver & 6108 & 6949 \\
Environment actions, EvoBelief & 2026 & 2737 \\
Environment actions, ReAct & 3839 & 3540 \\
Environment actions, WorldEvolver & 3051 & 3472 \\
\bottomrule
\end{tabular}
\end{table}

\begin{table}[t]
\centering
\footnotesize
\setlength{\tabcolsep}{3.5pt}
\caption{\textbf{The law validated on each fresh task instance.} $f$ is the reactor resonance in Hz; $c_5$, $d$, $t$, and $q$ are the crystal's spectrometer channel 5, density, temperature, and quantum gap. $x$ is the held-out crystal's reading, Predicted the prediction frozen before the reactor was dialed, and Dialed the frequency at which the reactor then activated. Rivals is the number of rival laws falsified on that task instance, Plan the outcome of the plan derived from the law for the target crystal (a passed plan completes the task), and Knowledge the judge's score on that task instance. Normal seeds 13 and 18 and Challenge seeds 9, 14, and 19 hide the variable in a radiation reading with random background; on Challenge seed 20 no rival survived within budget, and on seed 17 the law was validated but its plan never settled.}
\label{tab:dw-laws}
\begin{tabular}{@{}llrrrrlr@{}}
\toprule
Seed & Validated law & $x$ & Predicted & Dialed & Rivals & Plan & Knowledge \\
\midrule
\multicolumn{8}{@{}l}{\textit{Normal}} \\
s9 & $f = 104.99\,c_5 + 107.25$ & 30.50 & 3309.50 & 3310.00 & 18 & pass & 1.00 \\
s10 & $f = 99.01\,d + 102.24$ & 61.31 & 6172.69 & 6173.00 & 14 & pass & 1.00 \\
s11 & $f = 105.00\,t + 95.96$ & 17.81 & 1966.05 & 1966.05 & 18 & pass & 1.00 \\
s12 & $f = 105.01\,q + 90.40$ & 49.72 & 5311.60 & 5312.00 & 16 & pass & 1.00 \\
s13 & none (radiation seed) &  &  &  & 7 & -- & 0.00 \\
s14 & $f = 102.99\,c_5 + 91.92$ & 39.09 & 4118.27 & 4118.27 & 12 & pass & 1.00 \\
s15 & $f = 108.00\,d + 99.69$ & 34.31 & 3805.48 & 3805.48 & 12 & pass & 1.00 \\
s16 & $f = 106.01\,t + 102.64$ & 46.75 & 5058.50 & 5059.00 & 18 & pass & 1.00 \\
s17 & $f = 96.00\,q + 90.93$ & 69.78 & 6789.88 & 6790.00 & 10 & pass & 0.50 \\
s18 & none (radiation seed) &  &  &  & 9 & -- & 0.00 \\
s19 & $f = 105.99\,c_5 + 98.23$ & 16.28 & 1823.68 & 1824.00 & 12 & pass & 1.00 \\
s26 & $f = 105.99\,t + 100.10$ & 48.04 & 5192.24 & 5192.24 & 11 & pass & 1.00 \\
\midrule
\multicolumn{8}{@{}l}{\textit{Challenge}} \\
s9 & none (radiation seed) &  &  &  & 4 & -- & 0.00 \\
s10 & $f = 14.00\,t^2 + 32.99\,t + 568.99$ & 11.50 & 2800.00 & 2800.00 & 17 & pass & 1.00 \\
s11 & $f = 17.00\,d^2 + 25.85\,d + 510.48$ & 15.61 & 5058.63 & 5059.00 & 21 & pass & 1.00 \\
s12 & $f = 17.01\,q^2 + 20.72\,q + 783.45$ & 5.49 & 1409.67 & 1410.00 & 15 & pass & 1.00 \\
s13 & $f = 17.99\,c_5^2 + 29.04\,c_5 + 645.76$ & 12.84 & 3985.97 & 3986.00 & 18 & pass & 1.00 \\
s14 & none (radiation seed) &  &  &  & 7 & -- & 0.00 \\
s15 & $f = 19.00\,t^2 + 29.88\,t + 212.78$ & 18.57 & 7321.15 & 7321.15 & 15 & pass & 1.00 \\
s16 & $f = 17.99\,d^2 + 33.16\,d + 41.08$ & 11.85 & 2960.66 & 2961.00 & 5 & pass & 0.50 \\
s17 & $f = 12.89\,q^2 + 23.02\,q + 790.49$ & 7.12 & 1608.01 & 1608.55 & 9 & unsettled & 0.50 \\
s18 & $f = 14.99\,c_5^2 + 24.03\,c_5 + 581.88$ & 17.27 & 5470.28 & 5470.27 & 18 & pass & 1.00 \\
s19 & none (radiation seed) &  &  &  & 3 & -- & 0.00 \\
s20 & none &  &  &  & 7 & -- & 0.50 \\
\bottomrule
\end{tabular}
\end{table}

\paragraph{Plant Nutrients: protocol.} The theme's hidden rule is categorical. A first version of the Plant Nutrients instantiation was run on six fresh task instances under a pre-declared performance gate and failed it: it completed no task instance, although its rules had recovered the hidden nutrient rule on every Normal task instance and a table consistent with the hidden combination on every Challenge task instance. The failure had three named causes. A correct rule was retired by a counterexample that was not one, because growth also requires a seed in the tile and seeds inside tiles are not published; the validated rule set the soil controller but did not carry the rest of the planting chain, which the actor took without the rule; and the knowledge surface described what appeared on the soil rather than what a planted seed would do, which the judge did not credit. The version reported in Table~\ref{tab:discoveryworld-results} is a revised implementation that addresses those three failure modes. A counterexample by absence is admitted only after the plan has publicly created the causal opportunity, that is, after the deposit, the refill, and a pre-declared number of maturation ticks; a validated rule controls the whole plan, from configuring the controller through binding, acquiring and depositing the seed, waiting for maturation, and observing; and the knowledge surface renders the causal chain and is scoped to the task instance. Development used only task instances previously used in the pilot. The code was then frozen and run once on the twelve Normal and twelve Challenge task instances of a fresh set that had been enumerated blind before the pilot ran and never executed by any method. One repair, the identification of the actuator on a previously used development Challenge instance, was implemented after the development checks passed and was re-validated offline and on one previously used development instance; the pilot is never pooled with the reported result.

\paragraph{Plant Nutrients: mechanism.} The DW Plant column of Table~\ref{tab:mechanism-summary} counts what happened on the twenty-four fresh task instances. Of the nineteen completed task instances, seventeen completed through a validated task plan; one completed through a discriminating plan whose task plan fell short, and one completed with no plan. The retirements are informative: fifteen committed rules were retired by counterexamples, nine of them after the plan itself had created the opportunity, and thirty-five absence observations that would have counted as counterexamples in the first version were deferred because the opportunity had not yet been created. Knowledge margins over the baselines, in percentage points with paired bootstrap intervals: on Normal, 58.3 [33.3, 83.3] over ReAct, 66.7 [25.0, 100.0] over WorldEvolver, and 75.0 [50.0, 100.0] over the Hypothesizer; on Challenge, 33.3 [8.3, 58.3] over ReAct, WorldEvolver, and the Hypothesizer. Completion contrasts are in Table~\ref{tab:paired-stats}.

\paragraph{The Hypothesizer baseline at two output ceilings.} The Hypothesizer consolidates its knowledge into one long structured reply per step. Under the output ceiling frozen for the original campaigns, 3,072 tokens, that reply was truncated on both Challenge tiers, three bounded resends failed the same way, and the predeclared attempt policy classifies a failure at a frozen input or output admission as a method failure that is not retried: the Reactor Lab run sealed two Challenge task instances and the Plant Nutrients run one. Table~\ref{tab:dw-hyp-original} reports those original runs. For a complete baseline we reran the full Hypothesizer evaluation under a separately fixed configuration whose only change is that ceiling, raised to 8,192 tokens, on the same task instances by hash, one attempt per cohort, blind until every cohort had sealed. Prompt, policy, backbone, budgets, populations, and judge were unchanged, and no other method was re-run. Under the raised ceiling eleven consolidation replies exceeded the old ceiling and none approached the new one, so each of them would have been a truncation before. Those re-runs are the Hypothesizer rows of Table~\ref{tab:discoveryworld-results}. The two Normal re-runs also show the run-to-run variation of the backbone: the Plant Nutrients completion count moved from eight to six on the same twelve task instances with the same prompt.

\begin{table}[t]
\centering
\small
\caption{\textbf{The Hypothesizer at its original 3,072-token output ceiling.} Same metrics and task instances as Table~\ref{tab:discoveryworld-results}, in percent; the Challenge runs ended on output truncation, and their means are over the task instances sealed before the failure.}
\label{tab:dw-hyp-original}
\begin{tabular}{@{}llrrrr@{}}
\toprule
Family & Tier & Sealed & Completion & Knowledge & Procedural \\
\midrule
Reactor Lab & Normal & 12 of 12 & 0.00 & 0.00 & 74.33 \\
Reactor Lab & Challenge & 2 of 12 & 0.00 & 0.00 & 74.14 \\
Plant Nutrients & Normal & 12 of 12 & 66.67 & 0.00 & 82.41 \\
Plant Nutrients & Challenge & 1 of 12 & 0.00 & 0.00 & 66.67 \\
\bottomrule
\end{tabular}
\end{table}

\subsection{Held-out Prediction Evaluation}

\paragraph{Held-out prediction evaluation.}
Table~\ref{tab:scienceworld_prediction} evaluates 1{,}500 transitions per benchmark from episodes disjoint from belief formation. On ScienceWorld the belief is formed on 90 development episodes, and on ALFWorld on the 140 seen-split development episodes, with the evaluated transitions drawn from the unseen split. Each transition is scored once, with Haiku 4.5 as the backbone for every row, one frozen prediction prompt, and greedy decoding. State-$\Delta$ F1 is the set F1 between predicted and realized canonical state-change atoms, averaged over transitions, with a correct prediction of no change scoring one; Feasibility is agreement on whether the action executes; EM and Token F1 compare one deterministic rendering of the predicted and the realized state record. Frozen WM is the controller without $W$; the WorldEvolver rows run the official implementation with its semantic memory alone and with both memories. The ALFWorld block is a post-hoc rescoring of sealed structured predictions under the ScienceWorld metric definitions, which were fixed before any ALFWorld aggregate under them was computed. Under a paired bootstrap over transitions (10{,}000 resamples), EvoBelief's lead over every other row on State-$\Delta$ F1, EM, and Token F1 has a 95\% interval excluding zero on both benchmarks; on Feasibility, WorldEvolver-full's lead on ScienceWorld excludes zero, while every ALFWorld difference has an interval that includes zero.

\begin{table}[t]
\centering
\small
\setlength{\tabcolsep}{3.5pt}
\caption{Held-out structured state-change prediction on ScienceWorld and ALFWorld, 1{,}500 transitions per benchmark, disjoint from belief formation. State-$\Delta$ F1 measures canonical state changes; Feasibility measures action executability; EM and Token F1 compare structured state. Haiku 4.5.}
\label{tab:scienceworld_prediction}
\begin{tabular}{llcccc}
\toprule
& Model
& State-$\Delta$ F1 $\uparrow$
& Feasibility $\uparrow$
& EM $\uparrow$
& Token F1 $\uparrow$ \\
\midrule
\multirow{4}{*}{ScienceWorld}
& Frozen WM (no evolving $W$) & 0.7258 & 0.7547 & 0.6013 & 0.7866 \\
& WorldEvolver (semantic-only) & 0.7061 & 0.7827 & 0.6027 & 0.7814 \\
& WorldEvolver (full) & 0.6680 & \textbf{0.8753} & 0.5993 & 0.7820 \\
& EvoBelief & \textbf{0.7530} & 0.8113 & \textbf{0.6407} & \textbf{0.8045} \\
\midrule
\multirow{4}{*}{ALFWorld}
& Frozen WM (no evolving $W$) & 0.7572 & 0.8453 & 0.6567 & 0.8080 \\
& WorldEvolver (semantic-only) & 0.6862 & 0.8500 & 0.5733 & 0.7607 \\
& WorldEvolver (full) & 0.7250 & 0.8400 & 0.6060 & 0.7849 \\
& EvoBelief & \textbf{0.8473} & \textbf{0.8513} & \textbf{0.7613} & \textbf{0.8698} \\
\bottomrule
\end{tabular}
\end{table}